\documentclass{article} 
\usepackage{iclr2027_conference,times}

\makeatletter
\let\iclrAnd\And
\let\iclrAND\AND
\makeatother

\usepackage{amsmath,amsfonts,bm}

\def\eqref#1{equation~\ref{#1}}
\def\Eqref#1{Equation~\ref{#1}}

\def\1{\bm{1}}

\DeclareMathAlphabet{\mathsfit}{\encodingdefault}{\sfdefault}{m}{sl}
\SetMathAlphabet{\mathsfit}{bold}{\encodingdefault}{\sfdefault}{bx}{n}

\usepackage{hyperref}
\usepackage{graphicx}
\usepackage{url}
\usepackage{wrapfig}
\usepackage{booktabs}
\usepackage{multirow}
\usepackage[table]{xcolor}
\usepackage{pifont}
\usepackage{fontawesome5}
\usepackage{array}
\usepackage{amsmath}
\usepackage{amssymb}
\usepackage{booktabs}
\usepackage{colortbl}
\usepackage{xcolor}
\usepackage{graphicx}
\usepackage{diagbox}
\usepackage{algorithm}
\usepackage{algpseudocode}
\usepackage[most]{tcolorbox}
\definecolor{xbest}{RGB}{166,214,170}      
\definecolor{xsecond}{RGB}{220,240,222}    

\definecolor{ybest}{RGB}{239,154,154}      
\definecolor{ysecond}{RGB}{252,220,220}    
\definecolor{meanbest}{RGB}{166,214,170}      
\definecolor{meansecond}{RGB}{220,240,222}    

\AtBeginDocument{\let\And\iclrAnd \let\AND\iclrAND}

\newcommand{\cmark}{\textcolor{green!60!black}{\checkmark}}
\newcommand{\xmark}{\textcolor{red}{\ding{55}}}

\title{WorldGuide: Learning Success–Failure Boundaries in Latent World Models for Vision-Language-Action Policies}

\author{
Lin Liu \\ School of Information and Communication Engineering \\
Dalian University of Technology \& Beta Infinity
\And
Lu Zhang\thanks{Corresponding author.}, Yunzhi Zhuge \& Huchuan Lu \\
School of Information and Communication, \\
Engineering Dalian University of Technology
\And
Wu Yang, Yuzheng Zhuang, \\
\textbf{Shuai Tao} \& \textbf{Wulong Liu} \\
Beta Infinity
\And
Ziying Song\thanks{Corresponding author.} \\
Nanyang Technological University
}

\iclrfinalcopy 
\begin{document}

\maketitle

\begin{abstract}
Latent world models offer a promising way to improve Vision-Language-Action policies by capturing the consequences of actions. However, models trained primarily on expert demonstrations have limited exposure to failure outcomes and may struggle to distinguish visually similar successful and failed interactions. 
We propose \textbf{WorldGuide}, a framework that learns these distinctions in latent space and uses them to guide policy training. WorldGuide combines predictive pretraining on successful and failed trajectories with contrastive learning on matched success--failure pairs. The learned predictor then provides a differentiable reward to guide joint optimization of the policy and visual encoder. The predictor is discarded after training, so deployment requires no additional world-model inference. Extensive experiments show that WorldGuide substantially improves VLA reliability and achieves state of the art performance on LIBERO 100 and SimplerEnv, reaching \textbf{96.8\%} and \textbf{72.0\%}, respectively. Code will be publicly available.
\end{abstract}

\section{Introduction}

Vision-Language-Action (VLA) models~\cite{pi_05,lingbot_va,gr00t} translate visual observations and language instructions into robot actions, providing a unified framework for learning manipulation skills from demonstrations. While imitation learning directly supervises policies to reproduce expert actions, it offers limited explicit supervision about how those actions affect subsequent states. World models~\cite{} complement this action supervision by predicting how interactions evolve. Recent latent world model approaches~\cite{univla,worldvla,Vla-jepa,flare} introduce future-representation prediction as an auxiliary objective, encouraging policies to encode action-relevant state transitions and thereby improving manipulation performance. 

\begin{figure*}[t]
\centering
\includegraphics[width=\textwidth]{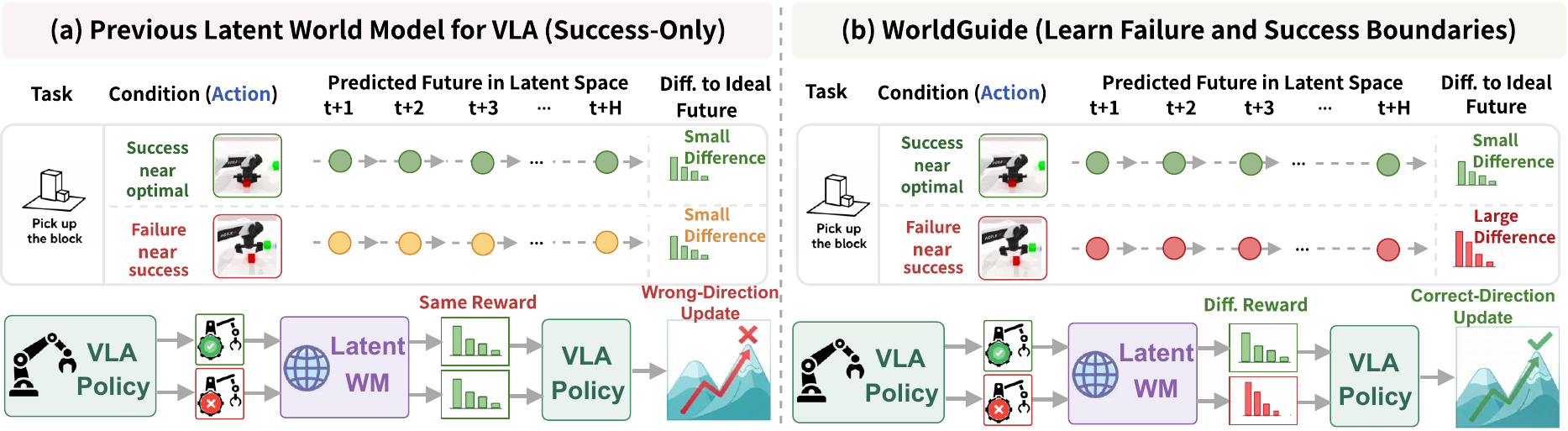}
\vspace{-5mm}
\caption{\textbf{Schematic motivation for WorldGuide.} (a) Success-only training poorly constrains latent dynamics for failures. (b) WorldGuide combines failure-rich predictive learning with matched comparisons to yield outcome-sensitive representations for policy guidance.}
\vspace{-3mm}
\label{fig:motivation}
\end{figure*}

Using latent predictions to guide policy improvement, however, requires the world model to capture the consequences of actions proposed by the policy, including those that depart from expert behavior. Consider a policy that closes its gripper slightly away from an object and then attempts to lift it. Although the action sequence resembles a successful grasp, the object remains on the table. A model trained predominantly on successful demonstrations receives limited supervision for this outcome and may still predict a future resembling successful execution. As illustrated in Fig.~\ref{fig:motivation}~(a), comparing such predictions with a successful goal can yield similar feedback for actions with different outcomes, providing little guidance for avoiding the failed interaction.

Training on failure trajectories expands this coverage by exposing the world model to unsuccessful transitions. Yet observing these transitions does not ensure that their outcome-relevant differences are prominent in the learned representation. A future-feature prediction objective does not explicitly prioritize differences according to their importance for task completion. Small discrepancies in contact or alignment may therefore contribute little to the latent distance even when they change the outcome. This motivates complementing predictive learning with supervision that makes successful and failed interactions more distinguishable.

Outcome labels provide such supervision, but how the corresponding trajectories are compared matters. Arbitrary successful and failed executions can differ in scene configuration, task progress, and motion patterns, allowing their representations to be separated using incidental cues. We therefore construct within-task comparisons between trajectories that are visually similar segments from the same task, matched around annotated failure onsets. A contrastive objective uses these matched examples to encourage outcome-discriminative representations, while the predictive objective preserves supervision for their state transitions. Together, these objectives address two complementary requirements: learning what happens during unsuccessful interactions and retaining the subtle distinctions that separate them from successful execution.


We introduce \textbf{WorldGuide}, which learns success--failure boundaries to guide policy optimization through three training stages. First, Failure-Rich Pretraining (Stage 1) mixes expert demonstrations with diverse failure trajectories, ensuring the world model learns complete physical dynamics rather than just success cases. Second, Boundary-Aware Contrastive Finetuning (Stage 2) pairs success and failure trajectories from the same task that look nearly identical in observations and actions. This forces the model to learn the true boundary between success and failure instead of relying on superficial shortcuts. Third, End-to-End Optimization with Differentiable Reward (Stage 3) freezes the trained world model and uses it as a penalty. When a candidate action is predicted to lead to a failed future, its prediction error is directly backpropagated to update the action head and visual encoder, steering the policy away from mistakes with zero deployment cost. In summary, our main contributions are summarized as follows:

\begin{itemize}
    \item We study the limitations of learning latent predictive feedback from successful demonstrations and examine how failure data and matched success--failure supervision affect outcome discrimination.
    
    \item We propose \textit{WorldGuide}, a three-stage training recipe that turns failure knowledge into policy guidance through failure-rich predictive pretraining, contrastive fine-tuning on matched interactions, and differentiable reward-guided policy training.
    
    \item Extensive simulation and real-world experiments show that \textit{WorldGuide} achieves competitive performance with the current SOTA model WAM while setting a new SOTA among VLA baselines (achieving \textbf{96.8\%} on LIBERO-100). Crucially, \textit{WorldGuide} reduces inference latency by \textbf{2.2$\times$ to 19.9$\times$} with zero deployment overhead from the world model.
\end{itemize}


\section{Related Work}

\paragraph{Robot Manipulation Policies.}
Data scale and model capacity have reshaped manipulation policies. Visuomotor imitation evolved into generalist formulations as RT recast control as tokenized sequence modeling, a paradigm OpenVLA consolidated over cross-embodiment datasets. Subsequent works introduced generative action heads via flow matching ($\pi_{0}$, $\pi_{0.5}$) for continuous high-frequency control, pursued efficient inference via linear sequence models (RoboMamba~\cite{robomamba}), or scaled foundation policies to humanoids (GR00T~\cite{gr00t}). Parallelly, world-action models (e.g., UniVLA~\cite{univla}, Video Policy~\cite{video_policy}, UWM~\cite{uwm}, FLARE~\cite{flare}, Cosmos~\cite{cosmos}) couple action generation with explicit future state prediction for long-horizon planning. Despite their generality, these systems rely almost exclusively on successful demonstrations, optimizing to succeed more often rather than recognizing failure boundaries.

\paragraph{World Models for Vision-Language-Action Learning.}
World models are integrated into VLA learning via three main paradigms. First, as \emph{simulators}~\cite{World-vla-loop,World-Env,VLA-RFT,WMPO}: generative video models synthesize rollouts for policy post-training, though hallucinations can yield physically inconsistent futures and unreliable rewards. Second, via \emph{joint learning}~\cite{worldvla,univla}: state prediction is coupled with action prediction at the feature level to internalize dynamics, but requiring inference-time rollouts inflates latency. Third, via \emph{JEPA-style objectives}~\cite{Vla-jepa}: latent prediction error acts as a training-only reward, avoiding inference overhead. However, across all paradigms, models are trained strictly on successful data without capturing failure boundaries. WorldGuide departs from this by pretraining on failure-rich transitions, employing boundary-aware contrastive learning on success--failure pairs, and back-propagating boundary rewards directly into action and visual modules.

\section{WorldGuide}
\label{sec:method}

\begin{figure*}[t]
\centering
\includegraphics[width=\textwidth]{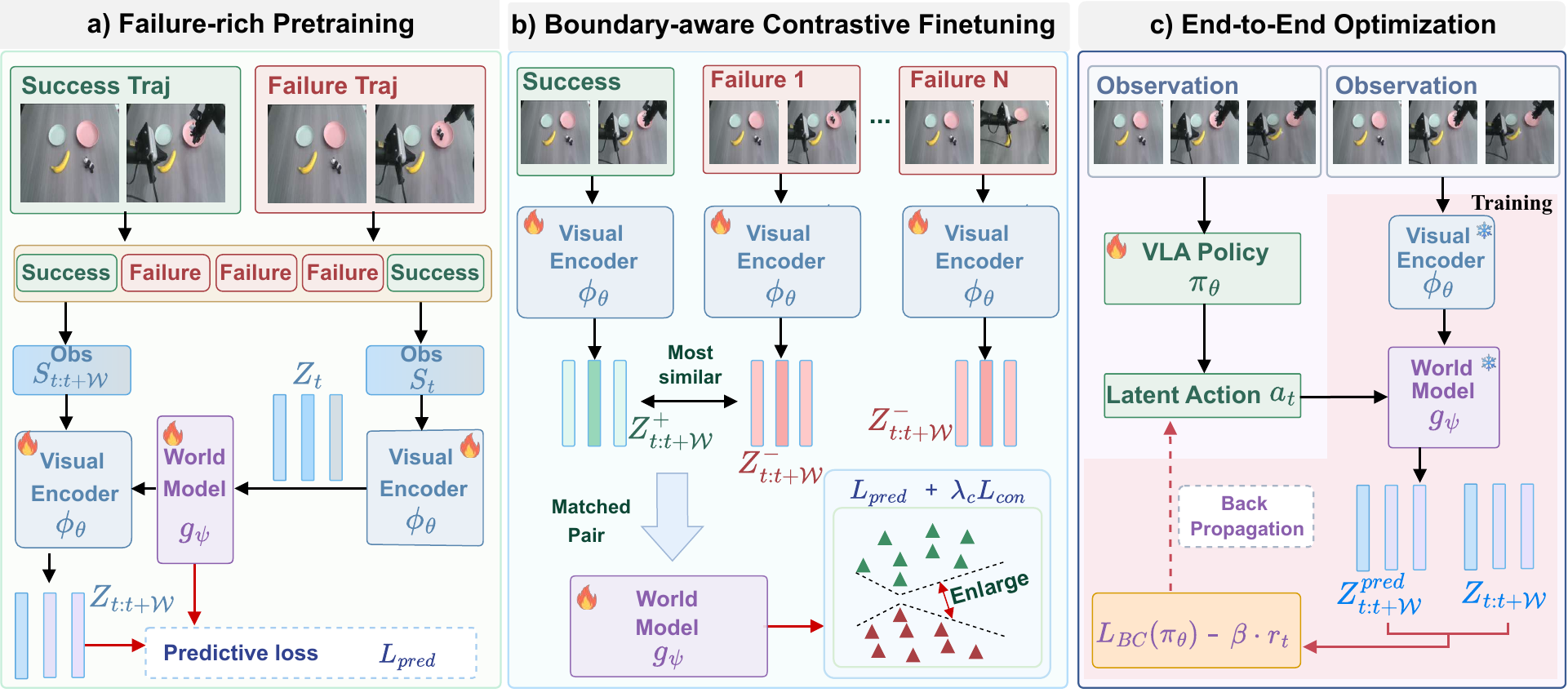}
\vspace{-5mm}
\caption{\textbf{Overview of WorldGuide.} \textbf{(a) Stage 1:} Failure-Rich Pretraining learns latent dynamics from successful and failed trajectories. \textbf{(b) Stage 2:} Boundary-Aware Contrastive Fine-Tuning distinguishes visually matched success--failure pairs while retaining the predictive objective. \textbf{(c) Stage 3:} End-to-End Optimization uses a differentiable reward from the frozen world model to train the VLA policy. Flame and snowflake icons indicate trainable and frozen modules, respectively.}
\vspace{-4mm}
\label{fig:overview}
\end{figure*}


As illustrated in Fig.~\ref{fig:overview}, WorldGuide comprises three sequential stages: (1) Stage 1 pretrains the visual encoder and latent world model on a failure-rich corpus to capture complete physical dynamics; (2) Stage 2 applies a boundary-aware contrastive objective on matched success--failure pairs to emphasize outcome-decisive execution differences; and (3) Stage 3 freezes the world model to optimize the VLA policy end-to-end via gradient backpropagation, completely discarding the world model at deployment for zero inference-time overhead.

\subsection{Preliminaries}
\label{sec:prelim}

\paragraph{Vision-language-action policy.}
A VLA policy $\pi_\theta$ maps a visual observation $s_{t}$ and a task instruction $lang$ to an action chunk $\hat{a}_{t:t+\mathcal{W}} = \pi_\theta(s_{t}, lang)$, where $\mathcal{W}$ denotes the observation and action horizons. In standard imitation learning, policies are trained on expert demonstrations $\mathcal{D}^{+} = \{\xi_i\}_{i=1}^{N}$, where each trajectory $\xi = \{(s_t, a_t)\}_{t=1}^{|\xi|}$ is annotated with instruction $lang$ and succeeds in completing the task (pred action: $\hat{a}_{t:t+\mathcal{W}}$, gt action: $a_{t:t+\mathcal{W}}$). To incorporate predictive dynamics, the visual encoder $\phi_\theta$ can be shared with an auxiliary world model, encouraging the policy's latent space to capture action-conditioned state transitions.

\paragraph{Latent World Model and Success Bias.}
Following the JEPA paradigm~\cite{Vla-jepa}, the latent world model $g_\psi$ predicts future semantic features rather than high-dimensional pixels:
\begin{equation}
z_{t:t+\mathcal{W}} = g_\psi\big(\phi_\theta(s_t),\, a_{t:t+\mathcal{W}}),
\label{eq:predict}
\end{equation}
where $\mathcal{W}$ denotes the prediction horizon. The predictor is trained via an $L_1$ prediction objective:
\begin{equation}
\mathcal{L}_{\mathrm{pred}}
=
\mathbb{E}_{\xi \sim \mathcal{D}}
\left[
\left\|
g_\psi\left(\phi_\theta(s_t), a_{t:t+\mathcal{W}}\right)
-
\phi_\theta(s_{t:t+\mathcal{W}})
\right\|_1
\right] + L_{reg},
\label{eq:pred_loss}
\end{equation}
where $\mathcal{L}_{\text{reg}}$ denotes the regularization loss employing the SIGReg regularizer (Details in Appendix~\ref{app:impl_train}). Ideally, $g_\psi$ measures alignment with the goal state via the latent distance between predicted future states and a target goal feature $z_{t:t+\mathcal{W}}^{\star} = \phi_\theta(s_{t:t+\mathcal{W}})$. However, when trained purely on expert demonstrations $\mathcal{D}^{+}$ (failure $\mathcal{D}^{-}$), the predictor suffers from \textbf{success bias}: its learned transition distribution collapses onto the success manifold. Consequently, failure states are erroneously mapped near successful latents, yielding optimistically low distance penalties even after execution errors: a failure mode that motivates our three stages framework. \textbf{Note: $t$ denotes the global timestep of the action chunk within the full trajectory, while $\tau$ represents the local frame index within the future action chunk}.

\subsection{Stage 1: Failure-Rich Pretraining}
\label{sec:stage1}

To capture a complete transition distribution, the world model must observe non-expert executions. We curate a failure dataset $\mathcal{D}^{-}$ from two complementary sources. First, \emph{simulator perturbations} create failures through noisy control, randomized object placements, and scripted errors. Preserving these failing episodes supplies \emph{breadth}, teaching the model coarse, low-level mistakes, like knocked-over objects or missed grasps. Second, \emph{intermediate checkpoints} saved during VLA policy training supply \emph{depth}. Their failures arise at critical decision points such as missed grasps or premature releases, producing subtle visual discrepancies that directly dictate task outcomes and densely cover the success--failure boundary. We joint pretrain $g_\psi$ and $\phi_\theta$ on the enriched corpus $\mathcal{D}^{+} \cup \mathcal{D}^{-}$ using the predictive loss in~\Eqref{eq:pred_loss}. Modeling this failure-rich dynamics forces $g_\psi$ to capture erroneous state transitions than snapping predictions back to the success manifold. As a result, the Stage~1 world model reliably penalizes coarse execution errors, setting the stage for Stage~2 to sharpen its resolution on near-boundary failures. More details can be found in Appendix.

\subsection{Stage 2: Boundary-Aware Contrastive Fine-Tuning}
\label{sec:stage2}

\paragraph{Progress-anchored pairing.}

\begin{figure*}[t]
\centering
\includegraphics[width=\textwidth]{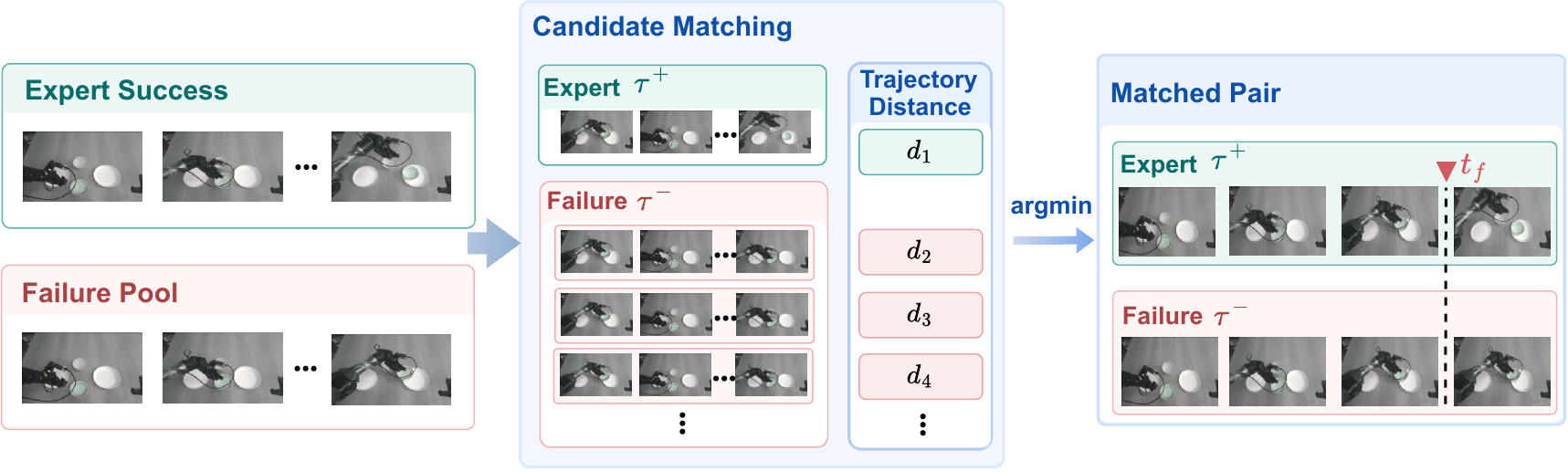}
\vspace{-5mm}
\caption{\textbf{Success--failure matching for contrastive learning.} Each successful segment is paired with the visually closest failure candidate from the same task within a temporal window to provide contrastive supervision; a dashed line marks the failure onset $t_f$.}
\vspace{-4mm}
\label{fig:pipeline}
\end{figure*}

As shown in Fig.~\ref{fig:pipeline}, Stage 2 builds each success failure pair by restricting candidates using a temporal proximity criterion before selecting a counterpart in feature space. Since failure rollouts share initial execution with successful demonstrations, step index $s$ serves as a unified task phase. The annotated failure onset $t_f$ (Appendix~\ref{app:vlm_annot}) anchors candidates within tolerance $\Delta = \mathcal{W}/2$ ($|t - t_f| \le \Delta$). Among valid candidates, the failure counterpart $\xi_i^-$ is chosen by feature distance over action chunk window $\mathcal{W_{t}}()$:
\begin{equation}
\xi_i^- = \operatorname*{arg\,min}_{\substack{\xi^- \in \mathcal{D}_{l_i}^- \\ \big| t - t_f \big| \le \Delta}} \frac{1}{|\mathcal{W}|} \sum_{\tau \in \mathcal{W}} \Big\| \phi_\theta\big(s_{t+\tau}(\xi_i^{+})\big) - \phi_\theta\big(s_{t+\tau}(\xi^-)\big) \Big\|_2^2,
\label{eq:match}
\end{equation}
where $\mathcal{D}_{l_i}^-$ is the failure set of task $l_i$, $\xi_i^{+}$ is the successful trajectory, and $\mathcal{W}_t(\xi)$ denotes the observation sequence window of length $\mathcal{W}$ centered at time $t$.
Let $\mathcal{H}$ denotes the \emph{hit set} comprising all valid success--failure pairs filtered via ~\Eqref{eq:match}. Segments without valid candidates ($i \notin \mathcal{H}$) are trained solely under $\mathcal{L}_{\mathrm{pred}}$. The selected counterpart is separated using loss weighted distance $d_m$ to satisfy ~\Eqref{eq:weighted_metric}. Finally, failure rollouts are truncated at post failure retry loops (Appendix~\ref{app:retry}), preventing repeated grasp attempts from dominating training.

\paragraph{Loss-weighted contrastive objective.}
Having matched pairs in the observation space, Stage~2 separates them within the latent space to focus gradients on task-decisive transitions. For each matched pair $i \in \mathcal{H}$, both trajectories are propagated through the encoder $\phi_\theta$ and the predictor $g_\psi$ over the window $\mathcal{W}$, yielding predicted future spatial-temporal latent tensors $z_{t+\mathcal{W}}^{+}, z_{t+\mathcal{W}}^{-} \in \mathbb{R}^{|\mathcal{W}| \times H \times W \times d}$. To evaluate their divergence without letting background noise dominate, each spatial-temporal tensor is summarized by pooling across both temporal and spatial dimensions into a channel descriptor $u = \frac{1}{|\mathcal{W}|HW} \sum_{\tau, h, w} z_{t+\tau, h, w} \in \mathbb{R}^d$. We then compute the loss-weighted distance $d_m\big(z_{t+\mathcal{W}}^{+}, z_{t+\mathcal{W}}^{-}\big)$:
\begin{equation}
d_m\big(z_{t+\mathcal{W}}^{+}, z_{t+\mathcal{W}}^{-}\big) = \sum_{\tau \in \mathcal{W}} \alpha_{\tau} \cdot \Big\| g \odot \big(z_{t+\tau}^{+} - z_{t+\tau}^{-}\big) \Big\|_F^2,
\label{eq:weighted_metric}
\end{equation}
\begin{equation}
\text{where} \quad
\alpha = \operatorname{softmax}\Big( \frac{1}{L H_a Q}\sum_{l=1}^L \sum_{k=1}^{H_a} \sum_{q=1}^Q A^{(l,k)}_{q, \cdot} \Big) \in \Delta^{|\mathcal{W}|},
\qquad
g = \sigma(W_g u) \in (0,1)^d.
\label{eq:gating_def}
\end{equation}
Here, $\odot$ denotes channel-wise broadcasting over spatial dimensions $(H, W)$, and $\|\cdot\|_F^2$ sums over spatial positions and channels. Dynamically, the temporal salience profile $\alpha \in \Delta^{|\mathcal{W}|}$ aggregates the predictor's cross-attention maps $A^{(l,k)}$ across $L$ layers, $H_a$ heads, and $Q$ query tokens to highlight execution-critical frames, while the channel mask $g = \sigma(W_g u) \in (0,1)^d$ suppresses static context to emphasize outcome-decisive channels.

The distance $d_m$ enters a one-sided repulsive hinge loss over all matched pairs in the hit set $\mathcal{H}$:
\begin{equation}
\mathcal{L}_{\text{con}} = \frac{1}{|\mathcal{H}|} \sum_{i \in \mathcal{H}} \left[ m - d_m\big(z_{t:t+\mathcal{W}}^{+}, z_{t:t+\mathcal{W}}^{-}\big) \right]_{+},
\label{eq:hinge_loss}
\end{equation}
where $m = 0.7$ denotes the margin and $[\cdot]_{+} = \max(0, \cdot)$. Stage~2 finetunes the pretrained world model via:
\begin{equation}
\mathcal{L}_{\text{ft}} = \mathcal{L}_{\text{pred}} + \lambda_c \mathcal{L}_{\text{con}},
\label{eq:stage2_finetune}
\end{equation}
with $\lambda_c$ balancing the contrastive loss and is set to $0.11$. Consequently, minimizing Equation~\ref{eq:stage2_finetune} pushes matched pairs beyond the margin and establishes a sharp, explicit boundary in the latent space that responds acutely to critical failure modes, thereby providing the exact structured latent signals required by Stage~3.

\subsection{Stage 3: End-to-End Optimization with Differentiable Reward}
\label{sec:stage3}

We freeze $g_\psi$ as a differentiable reward. For each state $s_t$ from expert data and policy rollouts, with a goal frame $s_{t:t+\mathcal{W}}$ from a successful demonstration of the same task, the reward of an action chunk is
\begin{equation}
r_t
=
-\,d_m\Big(
g_\psi\left(\phi_\theta(s_t),\, \hat{a}_{t:t+\mathcal{W}}\right),
\phi_\theta(s_{t:t+\mathcal{W}})
\Big).
\label{eq:reward}
\end{equation}

The policy is then optimized end-to-end while freezing the world model ($\phi_{\theta}$ and $g_{\psi}$), where $\phi_{\theta}$ is distinct from the VLA's vision encoder. 
\begin{equation}
\mathcal{L} = \mathcal{L}_{\mathrm{BC}}(\pi_\theta) - \beta\, \mathbb{E}\big[r_t\big],
\label{eq:policy}
\end{equation}
where $\mathcal{L}_{\mathrm{BC}}$ is the imitation objective on expert data and $\beta$ balances exploration toward reward-optimal behavior, which is set to 0.1. Reward gradients backpropagate through the frozen world model into the action head and VLM: the former favors goal-directed action chunks over failure-prone ones, while the latter aligns representations with safety-critical semantics. At deployment, the world model is discarded, leaving the VLA policy to execute independently.

\section{Experiments}
\label{sec:experiments}


We comprehensively evaluate WorldGuide across three widely recognized simulation benchmarks: LIBERO, RoboTwin, and SimplerEnv as well as on a suite of real-world robotic manipulation tasks. Details are provided in Appendix.



\subsection{Simulation Setup and Baselines}
\label{sec:setup}

\paragraph{Experimental Setup.}
We evaluate on LIBERO~\cite{libero} (100 (Long), Spatial, Goal and Object, Franka Panda), RoboTwin~\cite{robotwin} (50 tasks, 14-DoF Aloha-AgileX), and SimplerEnv~\cite{simplenv} (WidowX and Google Robot). Stage 1 pretraining uses 2K/2K, 6.6K/5.61K, and 140K/17K successful/failed trajectories for 60K, 120K, and 100K steps, respectively, with RoboTwin using 12/50 tasks. Stage 2 runs for 10K/20K/17K steps, followed by expert-data fine-tuning for 50K/110K/100K steps (stage 3). Real-robot experiments use 300 expert and 60 failure trajectories, with 30K/10K Stage 1/2 steps and 40K fine-tuning steps. All experiments use 8 NVIDIA H20 GPUs. OpenVLA-OFT and GR00T~\cite{gr00t} serve as the primary baselines for LIBERO/RoboTwin and SimplerEnv, respectively. Further details are in the Appendix.

\definecolor{meanbest}{RGB}{166,214,170}    
\definecolor{meansecond}{RGB}{220,240,222}  

\definecolor{lightblue}{RGB}{220,234,244}
\definecolor{lightyellow}{RGB}{255,242,204}

\begin{table*}[t]
\centering
\setlength{\abovecaptionskip}{0pt} 
\caption{\textbf{Quantitative Evaluation on LIBERO and SimplerEnv Benchmarks.} Results are reported as percentages. Best and second best overall results are highlighted in dark green and light green, respectively.}
\label{tab:combined_libero_simplerenv}
\small
\setlength{\tabcolsep}{4.5pt}
\renewcommand{\arraystretch}{1.12}

\setlength{\aboverulesep}{0pt}
\setlength{\belowrulesep}{0pt}

\resizebox{\textwidth}{!}{
\begin{tabular}{lccccccc ccc}
\toprule
\rowcolor{lightblue} \multicolumn{11}{c}{\textbf{(a) LIBERO Benchmark}} \\
\midrule
\textbf{Method} & \multicolumn{2}{c}{\textbf{Category}} & \textbf{Model Size} & \textbf{Latency (ms)} & \textbf{Speedup} & \textbf{100} & \textbf{Goal} & \textbf{Object} & \textbf{Spatial} & \textbf{Average} \\
\midrule
OpenVLA-OFT~\cite{openvla} & \multicolumn{2}{c}{VLA} & 7B & 277 & -- & 94.5 & 97.9 & 98.4 & 97.6 & 97.1 \\
$\pi_{0.5}$~\cite{pi_05} & \multicolumn{2}{c}{VLA} & 3.5B & 220 & -- & 92.4 & 98.0 & 98.2 & \cellcolor{meanbest}\textbf{98.8} & 96.9 \\
GR00T-N1.6~\cite{gr00t} & \multicolumn{2}{c}{VLA} & 3.3B & 259 & -- & 94.4 & 97.5 & 98.5 & 97.7 & 97.0 \\
UniVLA~\cite{univla} & \multicolumn{2}{c}{Latent-VLA} & 7B & -- & -- & 92.0 & 95.6 & 96.8 & 96.5 & 95.2 \\
Mantis~\cite{mantis} & \multicolumn{2}{c}{Latent-VLA} & 5.8B & -- & -- & 94.2 & 94.4 & 99.2 & \cellcolor{meanbest}\textbf{98.8} & 96.7 \\
VLA-JEPA~\cite{Vla-jepa} & \multicolumn{2}{c}{Latent-VLA} & 3B & -- & -- & 95.8 & 97.2 & 99.6 & 96.2 & 97.2 \\
\hline
Cosmos-Policy~\cite{cosmos-policy} & \multicolumn{2}{c}{WAM} & 2.1B & 1413 & 1.0$\times$ & \cellcolor{meansecond}97.6 & \cellcolor{meanbest}\textbf{98.2} & \cellcolor{meanbest}\textbf{100.0} & 98.1 & \cellcolor{meanbest}\textbf{98.5} \\
LingBot-VA~\cite{lingbot_va} & \multicolumn{2}{c}{WAM} & 5.5B & 4482 & 1.0$\times$ & \cellcolor{meanbest}\textbf{98.5} & 97.2 & \cellcolor{meansecond}99.6 & \cellcolor{meansecond}98.5 & \cellcolor{meanbest}\textbf{98.5} \\
Fast-WAM~\cite{fast_wam} & \multicolumn{2}{c}{WAM} & 6B & 486 & 1.0$\times$ & 95.2 & 97.0 & \cellcolor{meanbest}\textbf{100.0} & 98.2 & 97.6 \\
\hline
\textbf{WorldGuide (Ours)} & \multicolumn{2}{c}{\textbf{VLA (SOTA)}} & \textbf{4B} \textit{(+0.5B Train)} & \textbf{225} & \textbf{2.2$\times$--19.9$\times$} & \cellcolor{meansecond}\textbf{96.8} & \cellcolor{meansecond}\textbf{98.6} & \cellcolor{meansecond}\textbf{99.6} & \cellcolor{meansecond}\textbf{98.6} & \cellcolor{meansecond}\textbf{98.4} \\

\midrule
\rowcolor{lightyellow} \multicolumn{11}{c}{\textbf{SimplerEnv Benchmark}} \\
\midrule
\textbf{Method} & \multicolumn{5}{c|}{\textbf{Google Robot}} & \multicolumn{5}{c}{\textbf{WidowX Robot}} \\
\cmidrule{2-6} \cmidrule{7-11}
& \textbf{Pick} & \textbf{Move} & \textbf{Drawer} & \multicolumn{1}{c}{\textbf{Place}} & \multicolumn{1}{c|}{\textbf{Average}} & \textbf{Spoon} & \textbf{Carrot} & \textbf{Block} & \textbf{Eggplant} & \textbf{Average} \\
\midrule
LAPA$^*$~\cite{lapa} & -- & -- & -- & -- & \multicolumn{1}{c|}{--} & \cellcolor{meansecond}70.8 & 45.8 & \cellcolor{meanbest}54.2 & 58.3 & \textbf{57.3} \\
villa-x~\cite{villa-x} & 81.7 & 55.4 & 38.4 & \multicolumn{1}{c}{4.2} & \multicolumn{1}{c|}{44.9} & 48.3 & 24.2 & 19.2 & 71.7 & 40.8 \\
UniVLA~\cite{univla} & -- & -- & -- & \multicolumn{1}{c}{--} & \multicolumn{1}{c|}{--} & -- & -- & -- & -- & 42.7 \\
RoboVLMs~\cite{robovlms} & 77.3 & 61.7 & 43.5 & \multicolumn{1}{c}{24.1} & \multicolumn{1}{c|}{51.7} & 45.8 & 20.8 & 4.2 & \cellcolor{meansecond}79.2 & 37.5 \\
GR00T N1~\cite{gr00t} & 0.7 & 1.9 & 2.9 & \multicolumn{1}{c}{0.0} & \multicolumn{1}{c|}{1.4} & 1.4 & 0.0 & 0.0 & 13.9 & 3.8 \\
MoTo~\cite{moto} & 74.0 & 60.4 & 43.1 & \multicolumn{1}{c}{--} & \multicolumn{1}{c|}{--} & -- & -- & -- & -- & -- \\
OpenVLA-OFT~\cite{openvla} & -- & -- & -- & \multicolumn{1}{c}{--} & \multicolumn{1}{c|}{--} & 34.2 & 30.0 & \cellcolor{meansecond}30.0 & 72.5 & 41.8 \\
$\pi_0$~\cite{pi_0} & 72.7 & 65.3 & 38.3 & -- &\multicolumn{1}{c|}{--} & 29.1 & 0 & 16.6 & 62.5 & 27.1 \\
$\pi_0$-Fast~\cite{pi0_fast} & 75.3 & \cellcolor{meansecond}67.5 & 42.9 & -- & \multicolumn{1}{c|}{--} & 29.1 & 21.9 & 10.8 & 66.7 & 48.3 \\
VLA-JEPA~\cite{Vla-jepa} & \cellcolor{meansecond}88.3 & 64.1 & \cellcolor{meanbest}59.3 & \cellcolor{meansecond}49.1 & \multicolumn{1}{c|}{\cellcolor{meansecond} 65.2} & \cellcolor{meansecond}75.0 & \cellcolor{meanbest}70.8 & 12.5 & 70.8 & \cellcolor{meansecond}57.3 \\
\midrule
\textbf{WorldGuide (Ours)} & \cellcolor{meanbest}\textbf{91.3} & \cellcolor{meanbest}\textbf{75.0} & \textbf{57.6} & \textbf{64.1} &  \multicolumn{1}{c|}{\cellcolor{meanbest} 72.0} & \textbf{84.0} & \textbf{68.0} & \textbf{18.8} & \cellcolor{meanbest}\textbf{88.5} & \cellcolor{meanbest}\textbf{64.8} \\
\bottomrule
\end{tabular}}
\label{tab:1}
\end{table*}

\subsection{Simulation Evaluation}
\label{sec:sim}

\paragraph{LIBERO.}
As shown in Tab.~\ref{tab:1}, WorldGuide sets a new SOTA among standard zero-overhead VLAs with a \textbf{98.4\%} overall average on LIBERO, including \textbf{96.8\%} on LIBERO-100, outperforming OpenVLA-OFT (97.1\%) and GR00T-N1.6 (97.0\%) by over $1.3\%$. Crucially, WorldGuide matches heavy Generative WAMs (e.g., LingBot-VA at 98.5\%) while reducing per-step latency to 225\,ms ($2.2\times\text{--}19.9\times$ speedup). By discarding the world model during deployment, our framework eliminates the prohibitive memory and sampling overhead inherent to generative rollouts. Compared to expert-only latent models like VLA-JEPA (97.2\%), WorldGuide's margin expands to $+2.4\%$ on LIBERO-Spatial ($98.6\%$) and $+1.0\%$ on LIBERO-100 ($96.8\%$), proving the necessity of boundary-aware contrastive supervision over failure rollouts.

\begin{table}[t]
\centering
\setlength{\abovecaptionskip}{0pt} 
\caption{Comparison on RoboTwin 2.0 across 12 tasks. Each entry reports the success rate under the clean and the domain-randomized setups as $x \,/\, y$.}
\label{tab:policy_comparison}
\scriptsize
\setlength{\tabcolsep}{3.5pt}
\renewcommand{\arraystretch}{1.05}

\resizebox{\textwidth}{!}{
\begin{tabular}{lcccccc}
\toprule
& \multicolumn{6}{c}{\textbf{Task}} \\
\cmidrule(lr){2-7}
\textbf{Method}
& \textbf{Press Stapler}
& \textbf{Move Playingcard Away}
& \textbf{Place Object Stand}
& \textbf{Place Container Plate}
& \textbf{Turn Switch}
& \textbf{Lift Pot} \\
\midrule

$\pi_{0.5}$
& 0.97 \,/\, 0.95
& 0.94 \,/\, 0.98
& 0.87 \,/\, 0.87
& 0.95 \,/\, 0.93
& \cellcolor{meansecond}0.62 \,/\, 0.69
& 0.99 \,/\, 0.99 \\

$\pi_{0}$-Fast
& 0.96 \,/\, 0.97
& 0.95 \,/\, 0.98
& 0.86 \,/\, 0.92
& \cellcolor{meansecond}0.92 \,/\, 0.98
& 0.63 \,/\, 0.67
& \cellcolor{meansecond}1.00 \,/\, 0.99 \\

GR00T-N1.5
& \cellcolor{meansecond}0.98 \,/\, 0.98
& 0.99 \,/\, 0.99
& \cellcolor{meanbest}0.97 \,/\, 0.96
& 0.85 \,/\, 0.89
& 0.58 \,/\, 0.64
& 0.99 \,/\, 1.00 \\

StarVLA-OFT
& 0.97 \,/\, 0.96
& \cellcolor{meansecond}1.00 \,/\, 0.98
& 0.84 \,/\, 0.85
& 0.90 \,/\, 0.97
& 0.67 \,/\, 0.60
& \cellcolor{meanbest}1.00 \,/\, 1.00 \\

\textbf{WorldGuide}
& \cellcolor{meanbest}1.00 \,/\, 0.97
& \cellcolor{meanbest}0.99 \,/\, 1.00
& \cellcolor{meansecond}0.94 \,/\, 0.93
& \cellcolor{meanbest}0.99 \,/\, 0.97
& \cellcolor{meanbest} 0.66 \,/\, 0.69
& 0.99 \,/\, 1.00 \\

\bottomrule
\end{tabular}
}

\resizebox{\textwidth}{!}{
\begin{tabular}{lcccccc}
\toprule
\textbf{Method}
& \textbf{Adjust Bottle}
& \textbf{Place Phone Stand}
& \textbf{Place Mouse Pad}
& \textbf{Pick Diverse Bottles}
& \textbf{Rotate Qrcode}
& \textbf{Move Stapler Pad} \\
\midrule

$\pi_{0.5}$
& 0.95 \,/\, 0.91
& 0.51 \,/\, 0.66
& 0.66 \,/\, 0.57
& 0.55 \,/\, 0.51
& 0.59 \,/\, 0.59
& 0.39 \,/\, 0.41 \\

$\pi_{0}$-Fast
& 0.90 \,/\, 0.95
& 0.49 \,/\, 0.69
& 0.53 \,/\, 0.48
& \cellcolor{meansecond}0.65 \,/\, 0.63
& 0.50 \,/\, 0.61
& 0.33 \,/\, 0.39 \\

GR00T-N1.5
& 1.00 \,/\, 0.96
& \cellcolor{meansecond}0.92 \,/\, 0.98
& \cellcolor{meanbest}0.75 \,/\, 0.68
& 0.53 \,/\, 0.62
& \cellcolor{meanbest}0.87 \,/\, 0.87
& \cellcolor{meansecond}0.37 \,/\, 0.51 \\

StarVLA-OFT
& \cellcolor{meansecond}1.00 \,/\, 0.99
& 0.90 \,/\, 0.95
& 0.55 \,/\, 0.49
& 0.52 \,/\, 0.62
& 0.83 \,/\, 0.81
& 0.44 \,/\, 0.41 \\

\textbf{WorldGuide}
& \cellcolor{meanbest}1.00 \,/\, 1.00
& \cellcolor{meanbest}0.92 \,/\, 0.98
& \cellcolor{meansecond}0.77 \,/\, 0.73
& \cellcolor{meanbest}0.69 \,/\, 0.66
& \cellcolor{meansecond}0.89 \,/\, 0.85
& \cellcolor{meanbest}0.45 \,/\, 0.56 \\

\bottomrule
\end{tabular}
}

\end{table}

\paragraph{SimplerEnv.}
Relative to its closest method VLA-JEPA (Tab.~\ref{tab:1}), which shares a similar predictive paradigm but trains purely on expert demonstrations, WorldGuide exhibits clear superiority, outperforming VLA-JEPA by \textbf{+6.8\%} on Google Robot and \textbf{+7.5\%} on WidowX Robot. This gap highlights a key limitation of expert-only training: policies trained solely on successful executions cannot recognize the early signs of failure when perturbed by domain shifts. In contrast, our contrastive supervision explicitly penalizes trajectories near failure onsets ($t_f$), teaching the world model to recognize dangerous states.

\paragraph{RoboTwin.}
As shown in Tab.~\ref{tab:policy_comparison}, WorldGuide achieves the best average on RoboTwin 2.0 in both the clean setting at 85.7\% and the domain-randomized setting at 86.1\%, outperforming GR00T-N1.5 by 4.1 and 2.2 points and ranking among the top two on 8 of 12 tasks. The largest gain appears on Move-Stapler-Pad, where WorldGuide reaches 0.45/0.56 versus 0.44/0.41 for the baseline, a 15-point improvement under randomization. Turn-Switch is the exception, where fine-grained contact dominates and $\pi_{0.5}$ performs best. The learned boundary thus transfers to 14-DoF bimanual control and remains robust under strong domain randomization.

\subsection{Real-World Evaluation}
\label{sec:real}

We conduct real-world experiments on the ARX LIFT2 platform, collecting 300 human demonstrations covering three pick-and-place tasks of deliberately tight tolerances, so that small pose errors cascade into failure. $\pi_0$ and $\pi_{0.5}$ are fine-tuned on the same demonstrations and evaluated under identical settings, 20 trials per task per method. Further configurations are in the Appendix.

\begin{figure*}[t]
\centering
\includegraphics[width=\textwidth]{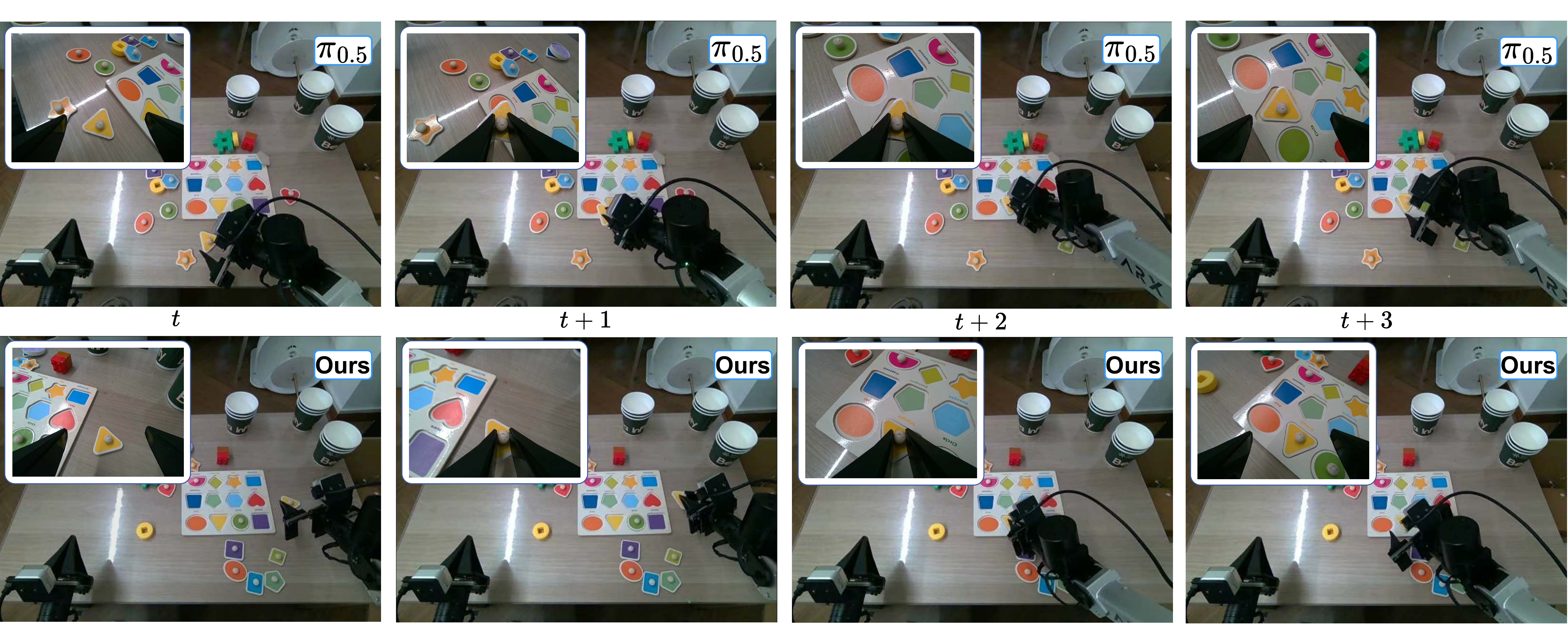}
\vspace{-6mm}
\setlength{\belowcaptionskip}{0pt}
\caption{Fine-grained manipulation comparison between $\pi_{0.5}$ and WorldGuide in real world.}
\label{fig:real_compare}
\end{figure*}

\begingroup
\setlength{\intextsep}{4pt}   
\setlength{\columnsep}{8pt}   

\begin{wraptable}{r}{0.52\textwidth}
\centering
\setlength{\abovecaptionskip}{0pt} 
\setlength{\tabcolsep}{2pt}
\caption{Real-world results on the ARX LIFT2. Success rate (\%) over \textbf{20 trials} per task per method}
\label{tab:real_exp}
\resizebox{\linewidth}{!}{
\begin{tabular}{ccccc}
\toprule
\diagbox{Method}{Task} & Purple square & Green circle & Yellow triangle & Avg.\ \\ \midrule
$\pi_{0}$       &         10\%              &          \textbf{20\%}            &       5\%              &     11.7\%          \\
$\pi_{0.5}$       &          10\%             &          15\%            &                    10\%     &   11.7\%        \\
WorldGuide       &          \textbf{20\%}             &         \textbf{20\%}             &      \textbf{15\%}        &   \textbf{18.3\%}             \\

\bottomrule
\end{tabular}}
\end{wraptable}

As shown in Tab.~\ref{tab:real_exp}, all three policies operate in the same low-success regime, with $\pi_0$ and $\pi_{0.5}$ both averaging 11.7\%, confirming that the difficulty is intrinsic to the tasks rather than any training budget. WorldGuide outperforms $\pi_{0.5}$ across all three tasks and beats or ties $\pi_0$, raising the pooled success rate to $18.3\%$ ($+57\%$ relative gain). Gains align directly with geometric sensitivity: vanishing on rotationally symmetric circles, doubling on squares, and tripling on triangles, which are hardest to grasp reliably. This distribution confirms that improvements concentrate at pose critical steps near the failure boundary rather than reflecting generic capacity gains (Fig.~\ref{fig:real_compare}). Given $20$ trials per task, the core signal lies in this consistent directional alignment rather than individual margins. Finally, all policies struggle with failure recovery, frequently repeating unsuccessful motions after initial errors.
\par
\endgroup

\subsection{Further Analysis and Ablation Study}
\label{sec:ablation}
%

\begin{figure*}[t]
\centering
\includegraphics[width=\textwidth]{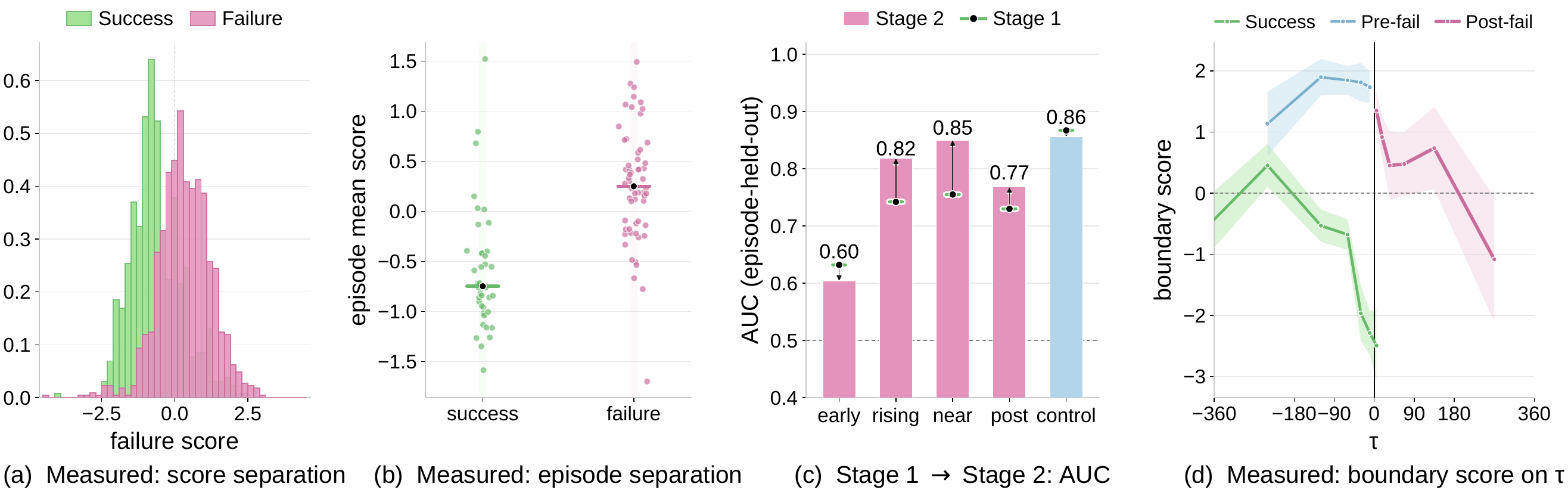}
\vspace{-5mm}
\setlength{\belowcaptionskip}{0pt}
\caption{Failure success separability of the world model predictor.}
\vspace{-4mm}
\label{fig:vis_sep}
\end{figure*}

\paragraph{Q1: Are Failure and Success Truly Separable in Feature Space?}
To evaluate feature separability, an episode-held-out linear readout (PCA + Logistic Regression) is fitted on predicted future embeddings across LIBERO (Fig.~\ref{fig:vis_sep} \& Tab.~\ref{tab:separability}).
%

The readout utilizes linear probes (logistic regression on world-model embeddings) trained and evaluated via episode-held-out cross-validation to ensure zero data leakage across episodes. The primary probe emits a signed \textcolor{red}{\emph{failure score}} (positive indicates failure). Single-window scores yield distinct distributions (Fig.~\ref{fig:vis_sep}a), while their per-episode average (\textcolor{red}{\emph{episode mean score}}) confirms trajectory-level separation beyond local window noise (Fig.~\ref{fig:vis_sep}b). Quantitatively, Stage~2 elevates episode-level AUC from $0.81$ to $0.87$ and critical-window AUC from $0.74$ to $0.79$. To characterize when the failure signature emerges relative to the annotated error, we train a second, \emph{phase-matched} probe to derive a \textcolor{red}{\emph{boundary score}}. Intuitively, this score measures whether a window resembles the moments immediately preceding an error (positive) or the winding-down phase of a success (negative). Formally, the probe is trained on late-stage failure windows ($\tau \in [-150, 0)$) and late success windows ($p \ge 0.70$), thereby decoupling elapsed time from classification. Projecting all windows, including the unseen post-fail segment ($\tau \ge 0$), onto this boundary score reveals that successful trajectories remain negative and deepen toward completion, whereas failing ones surge positive as $\tau \to 0$ (Fig.~\ref{fig:vis_sep}d). Before the error, Stage~2 separates impending failure from late success at $0.87$ AUC (vs.\ $0.78$ in Stage~1); beyond the error, this signature persists on held-out post-fail windows ($0.74$ AUC, with $81\%$ of episodes above the success median). Together, these results confirm that Stage~2 encodes a genuinely predictive failure signature rather than memorizing training artifacts.

\begingroup
\setlength{\intextsep}{4pt}   
\setlength{\columnsep}{8pt}   

\begin{wraptable}{r}{0.50\textwidth}
\centering
\setlength{\abovecaptionskip}{0pt} 
\caption{\textbf{Feature separability under Stage 2 boundary-aware fine-tuning.} Episode-held-out linear probing AUCs (Fig.~\ref{fig:vis_sep}) before (Stage~1) and after (Stage~2) fine-tuning. Middle rows detail AUC across task progress $p$ and failure offset $\tau$. Bottom rows evaluate early-warning generalization: training the probe strictly on near-failure frames ($\tau \in [-150, 0)$\, frames) and testing transferability to unseen future frames.}
\label{tab:separability}
\setlength{\tabcolsep}{5pt}
\renewcommand{\arraystretch}{1.08}
\resizebox{\linewidth}{!}{%
\begin{tabular}{lcc}
\toprule
\textbf{Readout / Evaluation Setting} & \textbf{Stage 1} & \textbf{Stage 2} \\
\midrule
Window AUC (decision-critical) & 0.74 & \textbf{0.79} \\
Episode AUC & 0.81 & \textbf{0.87} \\
\quad \textit{Phase:} Early ($p < 0.35$) & 0.63 & 0.60 \\
\quad \textit{Phase:} Mid ($0.35 \le p < 0.70$) & 0.74 & \textbf{0.82} \\
\quad \textit{Phase:} Near-failure ($p \ge 0.70$) & 0.75 & \textbf{0.85} \\
\quad \textit{Phase:} Post-failure ($\tau \ge 0$) & 0.73 & \textbf{0.77} \\
Boundary direction: Near-failure & 0.78 & \textbf{0.87} \\
Boundary direction: Unseen post-failure & 0.74 & \textbf{0.85} \\
\bottomrule
\end{tabular}%
}
\end{wraptable}
\paragraph{Q2: How much does the boundary-aware contrastive objective matter?}
Stage 2 improves all decision-critical readouts, window AUC from $0.74$ to $0.79$ and episode AUC from $0.81$ to $0.87$, yet the gains are uneven across phases: near-failure windows improve the most from $0.75$ to $0.85$, mid-phase windows follow from $0.74$ to $0.82$, and early-phase windows slightly drop from $0.63$ to $0.60$. The boundary-direction readout benefits most and generalizes furthest: fitted only on $\tau\in[-150,0)$, it rises from $0.78/0.74$ to $0.87/0.85$ on near-failure and never-trained post-failure windows, confirming that the contrastive objective concentrates representational capacity on the failure boundary rather than on generic progress or scene cues. Results from success only training further confirm the importance of failure-mixed training.
\par
\endgroup

\begingroup
\setlength{\intextsep}{4pt}   
\setlength{\columnsep}{8pt}   

\begin{wraptable}{r}{0.62\textwidth}
\centering
\setlength{\tabcolsep}{3pt}
\setlength{\abovecaptionskip}{0pt} 
\caption{\textbf{Stage-wise ablation.}}
\label{tab:stage_ablation}
\resizebox{\linewidth}{!}{%
\begin{tabular}{l*{7}{c}}
\toprule
\multirow{2}{*}{Setting} & \multicolumn{2}{c}{Stage} & \multicolumn{2}{c}{RoboTwin} & \multirow{2}{*}{LIBERO} & \multicolumn{2}{c}{SimplerEnv} \\
\cmidrule(lr){2-3}\cmidrule(lr){4-5}\cmidrule(lr){7-8}
 & 1 & 2 & Clean & Randomized & & Google & WidowX \\
\midrule
Base VLA & \xmark & \xmark & 80.1\% &80.2\% & 96.5\% & 68.8\% & 60.2\%\\
Stage 1 (success only) & \cmark & \xmark & 82.4\% & 81.9\% & 97.5\% & 69.6\% & 63.0\% \\
Stage 1 & \cmark & \xmark & 83.0\% & 83.5\% & 97.8\% & 70.7\% & 63.4\% \\
Stage 1 $+$ 2  & \cmark & \cmark & 85.7\% & 86.1\% & 98.4\% & 72.0\% & 64.8\% \\
\bottomrule
\end{tabular}}
\end{wraptable}

\paragraph{Q3: How much does each training stage contribute to policy performance?}

As shown in Tab.~\ref{tab:stage_ablation}, incorporating failure rollouts during Stage~1 pretraining outpaces training on success only data across all metrics, confirming that observing erroneous transitions builds a broader dynamics prior. Stage~2 contrastive finetuning yields consistent additional gains, particularly on challenging domains like RoboTwin and SimplerEnv, as the induced margin sharpens the success failure boundary in latent space to guide better policy decisions. Gains on LIBERO are more modest due to its near saturated baseline.
\par
\endgroup


\paragraph{Q4: What makes a world-model reward informative?}

Fig.~\ref{fig:reward_ablation} replaces only the Stage-3 reward under identical policy training. Goal similarity and prediction MSE from a success-only world model yield limited gains, reaching only $81.0/81.2\%$ and $82.5/82.8\%$ on RoboTwin, and their AUCs stay near chance at $0.47$--$0.50$ even in our failure-rich latent space, so the limitation lies in the reward readout itself. The boundary reward $-d_m()$ instead achieves an AUC of $0.85$ and improves RoboTwin to $85.7/86.1\%$ and LIBERO to $98.4\%$. Stage~3 is thus effective only when the reward can distinguish failure from success, which in turn depends on the failure-aware representation learned in Stages~1 and 2.

\begin{figure*}[h]
\centering
\includegraphics[width=\textwidth]{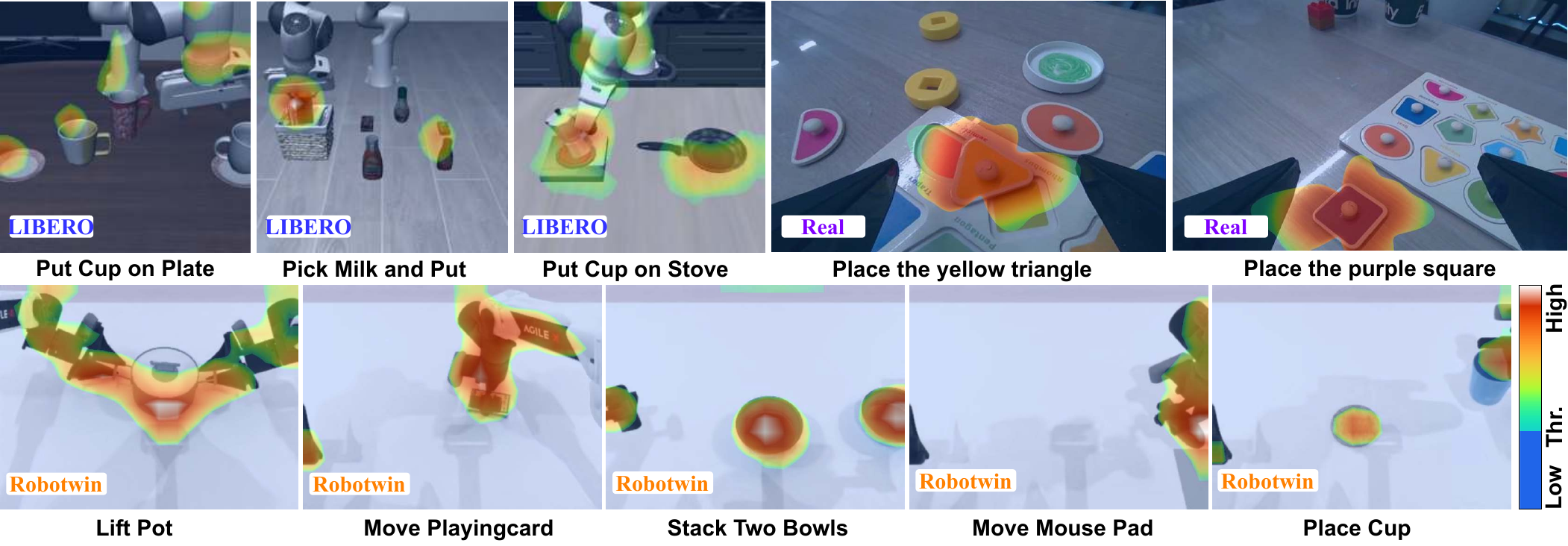}
\vspace{-5mm}
\setlength{\belowcaptionskip}{0pt}
\caption{Attention weight matrix of latent action tokens to image tokens.}
\vspace{-4mm}
\label{fig:attention_heatmap}
\end{figure*}

\begingroup
\setlength{\intextsep}{4pt}  
\setlength{\columnsep}{8pt}  

\begin{wrapfigure}{r}{0.60\textwidth}
\centering
\includegraphics[width=\linewidth]{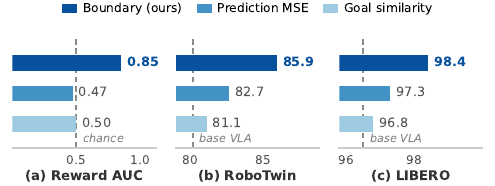}
\vspace{-5mm}
\setlength{\belowcaptionskip}{0pt}
\caption{\textbf{Reward-source ablation}.}
\label{fig:reward_ablation}
\end{wrapfigure}
\paragraph{Q5: How does world model reshape the representation?} Fig.~\ref{fig:attention_heatmap} shows the attention weights between action tokens and image patches, overlaid on the raw observations as a thresholded thermal map; only the top 3\%--10\% of attention mass is colored. The hotspot is not a static saliency prior but consistently lands on the manipulated object across scenes. This demonstrates that the attention mechanism dynamically grounds decision-making on task-relevant contact regions rather than background context.
\par
\endgroup

\section{Conclusion}

We introduced WorldGuide, a latent world model that enhances VLA robustness by explicitly modeling the boundary between successful and failed behaviors. Learning jointly from success and failure trajectories with boundary-aware contrastive learning yields failure-sensitive representations, and the world model is discarded at inference. Experiments across simulation and real-world tasks show improved robustness and task success with no additional inference overhead, highlighting the promise of failure-aware latent world modeling for embodied intelligence.

\subsection*{AI Use Statement}
In this work, generative AI tools were used to assist with method code implementation, language polishing, figure layout and color adjustments, as well as automatic temporal timestamp annotation for $t_f$. Specifically, the open source Qwen 3.5 model was applied to suggest timestamp annotations for transform frames, and all model generated annotations were manually sampled and cross verified by the authors to guarantee data quality. Generative AI was not used to develop theoretical models or conceptual frameworks, or to formulate or refine hypotheses. It was not used to provide essential elements for mathematical proofs, assist in writing proofs, or support qualitative or thematic data analysis.

All AI assisted outputs were thoroughly reviewed and verified by the authors. Any code or annotations generated with the assistance of large language models were independently tested and verified for correctness. The authors take full responsibility for the final content of this work, including all text, claims, data annotations, and materials produced with the assistance of generative AI. Please refer to the Appendix for further details.

\subsection*{Reproducibility Statement}
Section~\ref{sec:method} specifies the WorldGuide architectures, training objectives, and inference procedures. Section~\ref{sec:experiments} details the training datasets, benchmark splits, evaluation metrics, and comparison baseline settings. The Appendix~\ref{app:impl} provides full model configurations, optimization hyperparameters, evaluator definitions, and metric rules. Upon acceptance, we will publicly release all source code, evaluation protocols, and model checkpoints required to reproduce all reported results.

\bibliography{iclr2027_conference}
\bibliographystyle{iclr2027_conference}

\appendix

\newcommand{\figtodo}[1]{%
\begin{center}\fbox{\parbox{0.92\linewidth}{\vspace{14pt}\centering\textcolor{red}{TODO (figure to add): #1}\vspace{14pt}}}\end{center}}

\section{Implementation Details}
\label{app:impl}

\subsection{Architecture and Base Models}
WorldGuide is implemented on the StarVLA training framework. The policy backbone is Qwen3-VL-4B-Instruct; per benchmark we attach the standard action head of the corresponding base VLA (discrete OFT action tokens for LIBERO and RoboTwin, a DiT-B flow-matching head for SimplerEnv), so that ``Base VLA'' in all ablations refers to the same publicly comparable recipe fine-tuned on the same demonstrations. The world model $g_\psi$ follows the JEPA design: a ViT visual encoder $\phi_\theta$ shared with the policy and a lightweight transformer predictor that maps a window of past latents and actions to the predicted future latent. Tab.~\ref{tab:app_arch} summarizes the per-benchmark configurations.

\begin{table}[h]
\centering
\caption{Architecture and per-benchmark configurations. The world model is discarded at deployment; only the VLA policy executes.}
\label{tab:app_arch}
\small
\setlength{\tabcolsep}{6pt}
\begin{tabular}{lccc}
\toprule
 & LIBERO & RoboTwin 2.0 & SimplerEnv \\
\midrule
VLM backbone & Qwen3-VL-4B & Qwen3-VL-4B & Qwen3-VL-4B \\
Action head & OFT tokens & OFT tokens & DiT-B flow matching \\
Action space & $\Delta$qpos, 7-DoF & abs qpos, 14-DoF & $\Delta$EE, 7-DoF \\
Action chunk $H$ & 8 & 50 & 16 \\
WM encoder $\phi_\theta$ & ViT-L/14, $224^2$ & ViT-L/14, $224^2$ & ViT-T/14, $224^2$ \\
WM window (ctx $+$ pred) & $8+1$, frameskip 1 & $50+1$, frameskip 1 & $16+1$, frameskip 1 \\
WM predictor & 6-layer, 16-head & 6-layer, 16-head & 4-layer, 8-head \\
Latent dim & 768 & 768 & 768 \\
WM parameters & 0.5B & 0.5B & 0.5B \\
Deployment params & 4B (no WM) & 4B (no WM) & 4B (no WM) \\
\bottomrule
\end{tabular}
\end{table}

The world model predicts next-frame latents in parallel over sliding contexts via causal teacher-forcing, eliminating autoregressive rollouts during training. For clarity, we simplify the frame input at step $t$ to $s_t$. Within a temporal window, the output at position $t$ predicts the latent of frame $s_{t{+}1}$ conditioned on all preceding frames $\le s_t$, allowing a single forward pass to yield all next-frame predictions simultaneously—each supervised under its respective historical context. Regarding window configurations, the world model observes an $(8{+}1)$-frame sequence for LIBERO and a $(50{+}1)$-frame sequence for RoboTwin, matching the policy's observation range. For SimplerEnv, the data registry provides a $(16{+}17$-frame window aligned with the $16$-step action chunk, where the first frame conditions the VLM and the full $17$-frame sequence feeds into the world model. 

\subsection{Training Configuration}
\label{app:impl_train}
All stages share the same optimizer setup: AdamW with cosine learning-rate decay to a minimum floor, gradient clipping at a threshold of $1.0$, mixed-precision training (bf16), and gradient checkpointing distributed across 8$\times$ NVIDIA H20 GPUs. Tab.~\ref{tab:app_train} summarizes the stage-wise training configurations for LIBERO, RoboTwin, SimplerEnv, and real-robot evaluations, with detailed in Tab.~\ref{tab:app_arch} and Tab.~\ref{tab:app_train}.

\begin{table}[t]
\centering
\caption{Stage-wise training configurations across different environments.}
\label{tab:app_train}
\small
\setlength{\tabcolsep}{6pt}
\begin{tabular}{lccc}
\toprule
 & Stage 1 (pretrain) & Stage 2 (contrastive) & Stage 3 (reward) \\
\midrule
\multicolumn{4}{l}{\textbf{LIBERO}} \\
Steps & 60{,}000 & 10{,}000 & 50{,}000 \\
Batch / device & 16 & 16 & 16 \\
Learning rate & $5\times 10^{-5}$ (WM) & $5\times 10^{-5}$ (WM) & $1\times 10^{-5}$ (VLA) \\
LR schedule & cosine $\to 1\times 10^{-6}$ & cosine $\to 1\times 10^{-6}$ & cosine $\to 2\times 10^{-6}$ \\
Warmup ratio & 0.1 & 0.1 & 0.1 \\
Weight decay & $1\times 10^{-3}$ & $1\times 10^{-3}$ & $1\times 10^{-8}$ \\
Prediction loss $\mathcal{L}_{\mathrm{pred}}$ & $1.0$ & $1.0$ & -- \\
SIGReg regularizer & $0.09$ & $0.09$ & -- \\
Contrastive loss $\mathcal{L}_{\mathrm{con}}$ & -- & $0.11$ & -- \\
Reward weight $\beta$ & -- & -- & 0.1 \\
Initialization & Rand & Stage 1 WM & Stage 2 WM \\

\midrule
\multicolumn{4}{l}{\textbf{RoboTwin}} \\
Steps & 120{,}000 & 20{,}000 & 110{,}000 \\
Batch / device & 16 & 16 & 16 \\
Learning rate & $6.5\times 10^{-5}$ (WM) & $6.5\times 10^{-5}$ (WM) & $2\times 10^{-5}$ (VLA) \\
LR schedule & cosine $\to 2\times 10^{-6}$ & cosine $\to 2\times 10^{-6}$ & cosine $\to 2.5\times 10^{-6}$ \\
Warmup ratio & 0.1 & 0.1 & 0.1 \\
Weight decay & $1\times 10^{-3}$ & $1\times 10^{-3}$ & $1\times 10^{-8}$ \\
Prediction loss $\mathcal{L}_{\mathrm{pred}}$ & $1.0$ & $1.0$ & -- \\
SIGReg regularizer & $0.09$ & $0.09$ & -- \\
Contrastive loss $\mathcal{L}_{\mathrm{con}}$ & -- & $0.11$ & -- \\
Reward weight $\beta$ & -- & -- & 0.1 \\
Initialization & Rand & Stage 1 WM & Stage 2 WM \\

\midrule
\multicolumn{4}{l}{\textbf{SimplerEnv}} \\
Steps & 100{,}000 & 17{,}000 & 100{,}000 \\
Batch / device & 16 & 16 & 16 \\
Learning rate & $5\times 10^{-5}$ (WM) & $5\times 10^{-5}$ (WM) & $1\times 10^{-5}$ (VLA) \\
LR schedule & cosine $\to 1\times 10^{-6}$ & cosine $\to 1\times 10^{-6}$ & cosine $\to 1.5\times 10^{-6}$ \\
Warmup ratio & 0.1 & 0.1 & 0.1 \\
Weight decay & $1\times 10^{-3}$ & $1\times 10^{-3}$ & $1\times 10^{-8}$ \\
Prediction loss $\mathcal{L}_{\mathrm{pred}}$ & $1.0$ & $1.0$ & -- \\
SIGReg regularizer & $0.09$ & $0.09$ & -- \\
Contrastive loss $\mathcal{L}_{\mathrm{con}}$ & -- & $0.11$ & -- \\
Reward weight $\beta$ & -- & -- & 0.1 \\
Initialization & Rand & Stage 1 WM& Stage 2 WM \\

\midrule
\multicolumn{4}{l}{\textbf{Real Env}} \\
Steps & 30{,}000 & 10{,}000 & 40{,}000 \\
Batch / device & 24 & 24 & 24 \\
Learning rate & $2.5\times 10^{-5}$ (WM) & $2.5\times 10^{-5}$ (WM) & $2\times 10^{-5}$ (VLA) \\
LR schedule & cosine $\to 2.5\times 10^{-6}$ & cosine $\to 2.5\times 10^{-6}$ & cosine $\to 2.5\times 10^{-6}$ \\
Warmup ratio & 0.1 & 0.1 & 0.1 \\
Weight decay & $1\times 10^{-2}$ & $1\times 10^{-2}$ & $1\times 10^{-2}$ \\
Prediction loss $\mathcal{L}_{\mathrm{pred}}$ & $1.0$ & $1.0$ & -- \\
SIGReg regularizer & $0.09$ & $0.09$ & -- \\
Contrastive loss $\mathcal{L}_{\mathrm{con}}$ & -- & $0.11$ & -- \\
Reward weight $\beta$ & -- & -- & 0.1 \\
Initialization & Rand & Stage 1 WM& Stage 2 WM \\
\bottomrule
\end{tabular}
\end{table}

\paragraph{Stage-2 Hyperparameters.} The temporal match tolerance is set to half of action chunk size, with a hinge margin of $m = 0.7$ and a contrastive loss weight of $\lambda_c = 0.11$. Both the temporal salience profile $\alpha$ and the channel gate $g$ (\Eqref{eq:gating_def}) are enabled by default in all main experiments; their corresponding ablation variants (i.e., uniform $\alpha$ and $g \equiv \mathbf{1}$) are evaluated in Tab.~\ref{tab:contrastive_variants}. Candidate failure windows are indexed via a 64 entry LRU cache and encoded in a single batched forward pass (with gradients disabled) per step, ensuring that nearest-candidate selection incurs negligible computational overhead.

\paragraph{Regularization.} 
To prevent representation collapse during predictive loss minimization in Stages 1 and 2, we incorporate SIGReg, a VICReg-style variance--covariance regularizer acting directly on the predicted latent representations.Specifically, SIGReg encourages the latent distribution to match an isotropic standard Gaussian prior by matching their empirical characteristic functions over random projections. In our setup, we configure SIGReg with 17 evaluation knots and 1024 random projection directions, applying a weighting coefficient of $0.09$ across all environments. By penalizing feature cross-correlations and maintaining spatial variance across latent dimensions without requiring negative pairs or asymmetric architectures, this regularizer ensures that the world model produces informative, high entropy representations that effectively ground downstream reward modeling and policy learning.

\section{Evaluation Protocols and Real-World Setup}
\label{app:eval}

\begin{wrapfigure}{r}{0.50\textwidth}
    \centering
    \includegraphics[width=0.50\textwidth]{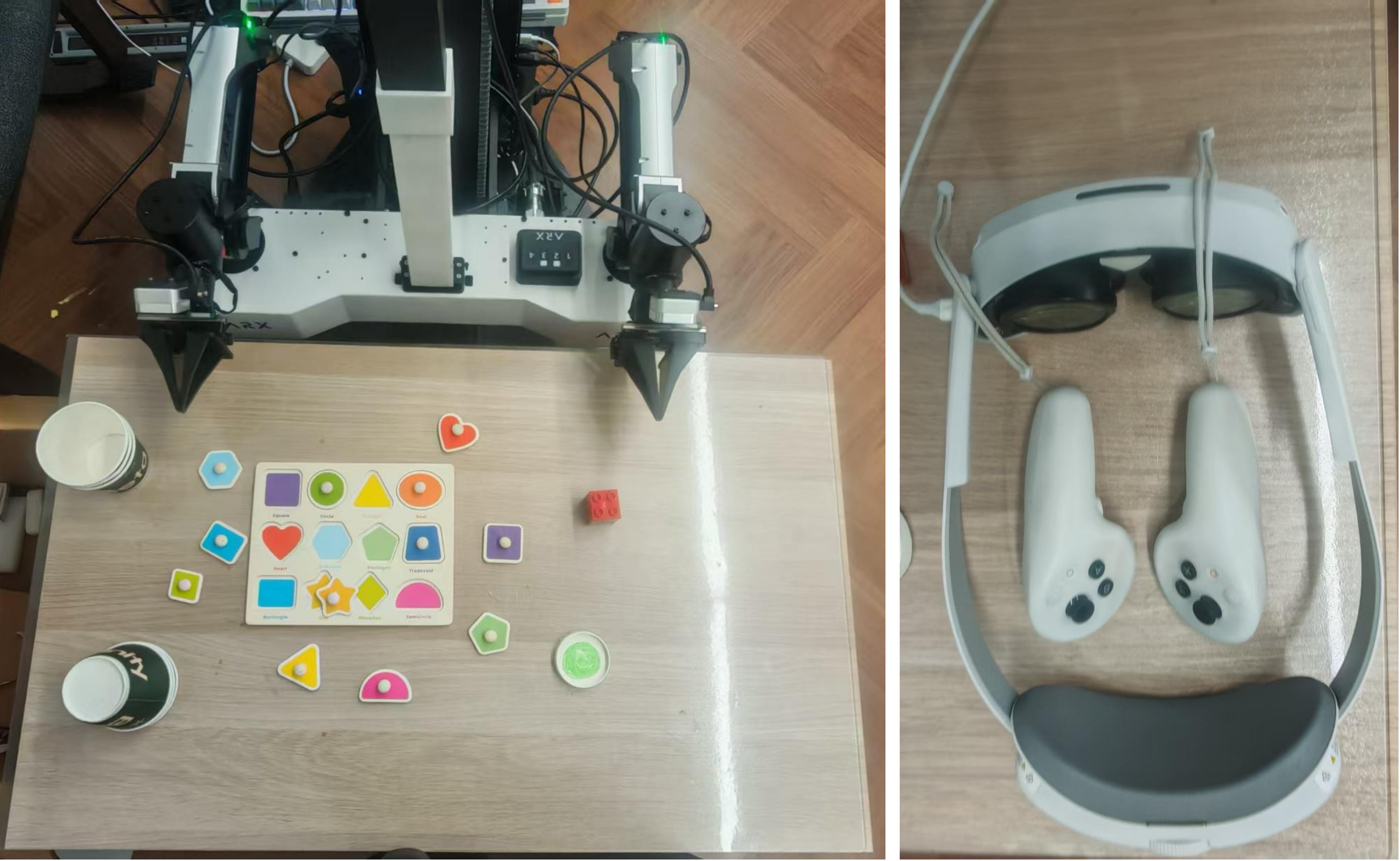}
    \caption{Real world experimental setup: ARX Lift 2 with PICO.}
    \label{fig:stats}
\end{wrapfigure}

\subsection{Simulation Benchmarks}
\textbf{LIBERO} covers four suites (Object, Goal, Spatial, and 100) average success rates. \textbf{RoboTwin 2.0} is evaluated over 12 tasks under both the clean and the domain-randomized settings (scene clutter, lighting, backgrounds, object configurations, and instruction paraphrases) on the 14-DoF Aloha-AgileX bimanual platform. Each cell in Tab.~\ref{tab:policy_comparison} is reported as clean/randomized. \textbf{SimplerEnv} evaluates real-to-sim visual generalization (lighting, textures, camera poses) on the Google Robot and WidowX embodiments. All policies are served through a WebSocket policy server (client/server split between the evaluation environment and the model host) and execute with the different action-chunk frequency (8 chunk sizes (LIBERO), 50 chunk sizes (Robotwin), 16 chunk sizes (SimplerEnv)). The latency in Tab.~\ref{tab:1} (225\,ms per chunk for WorldGuide) is measured on this deployment path, i.e., the world model contributes nothing at inference time.


\subsection{Real-World Setup}
Real robot experiments use the ARX LIFT2 dual arms platform teleoperated with a PICO headset. We collect 300 human demonstrations for three pick-and-place tasks with different object geometries (purple square, green circle, yellow triangle) and deliberately tight tolerances, so that small pose errors cascade into failure. $\pi_0$, $\pi_{0.5}$, and WorldGuide are fine-tuned on the same demonstrations and evaluated interleaved under identical lighting and object layouts, 20 trials per task per method, to control for environment drift. Since the aforementioned tasks can be completed using a single arm, we collect demonstrations using only one arm. During training and deployment, we mask out the data and actions associated with the irrelevant arm. For real-world model training, we use the open-source LeRobot framework.

\section{Failure Corpus Construction}
\label{app:failure}

\subsection{Two Complementary Sources}
The failure set $\mathcal{D}^{-}$ merges two sources, retained by an automatic task-completion check.

\textbf{Simulator perturbations (broad).} We execute tasks in the simulator with perturbed initial object poses, injected action-control noise, and scripted skill violations that disregard predefined conditions (e.g., grasping before alignment). The resulting failed action sequences and scene data are added to $\mathcal{D}^{-}$, covering coarse and visually salient errors that span the broader regions of the failure space.

\textbf{Under-converged models (deep)}. We execute intermediate checkpoints saved during standard VLA fine-tuning in the simulator and retain the resulting failure episodes. During real-world execution, these under-converged models often produce failures that are visually almost indistinguishable from successful executions, yet errors emerge at critical stages, such as missed grasps, incorrect object grasps, or premature gripper release.

\subsection{VLM Keyframe Annotation of $t_f$}
\label{app:vlm_annot}
For each failure event, we render the collected data into a video and use the prompt template below (the English rendering is shown here, while the native-language prompt is included chinese in the released code) to annotate the failure onset time $t_f$ with Qwen3.5-VL-27B over the entire video. Each event is assigned a failure onset timestamp $t_f$, which is subsequently used for the progress-anchored pairing in~\Eqref{eq:match}.

\begin{tcolorbox}[breakable, enhanced,
  colback=gray!3, colframe=black!40, boxrule=0.6pt, arc=1.2mm,
  left=7pt, right=7pt, top=5pt, bottom=5pt,
  title={\bfseries VLM keyframe-annotation prompt (English rendering)},
  colbacktitle=black!82, coltitle=white]
\small\ttfamily
You are an expert in robot-manipulation video analysis.\\[2pt]
Watch the entire video and analyze the arm's failed operations.\\[2pt]
For each manipulated object, determine exactly one timestamp:\\[2pt]
\quad fail: the earliest moment at which the robot's grasp, placement, pick, rotate and other manipulation fails.\\[2pt]
Requirements:\\
\quad -- Analyze the entire video.\\
\quad -- Return the time in seconds (s).\\
\quad -- Precision: one decimal place.\\
\quad -- Exactly one fail time per manipulated block.\\
\quad -- If a timestamp is mildly ambiguous, give the most reasonable estimate
       from the video content; do not return null.\\
\quad -- Do not output any explanation.\\
\quad -- Output JSON only.\\[2pt]
Output format (strict):\\
\quad \{ "fail": 2.0 \}
\end{tcolorbox}

The decoding pipeline is deliberately defensive: (i) generated text is truncated at the \texttt{</mthink>} tag, and the trailing answer is parsed as JSON; (ii) if JSON parsing fails, the first \texttt{{...}} block is extracted via regular expressions as a fallback; and (iii) events that remain unparsable are skipped rather than inferred. The validated timestamps are converted to frame indices at the corpus frame rate (20~fps). We then perform manual verification by reviewing a random sample of annotated videos.

\subsection{Failure-Onset-Based Truncation}
\label{app:retry}
Rollouts that continue after the first failure often enter prolonged post-failure retry loops, introducing redundant data that can interfere with training. We therefore truncate each unlabeled failure episode based on the annotated failure onset timestamp $t_f$, retaining two additional action chunks after $t_f$ before truncation.

\subsection{Matched Pairing for Stage 2}
\label{app:pairing}
Stage 2 performs online success-failure pairing using~\Eqref{eq:match}. At timestamp $t$, for each loaded successful segment of task $l$, we first collect \emph{all} annotated failure events of the same task whose failure onset satisfies $|t-t_f|\le\Delta$. A failure video is considered a candidate if the current timestamp $t$ falls within an action-chunk-sized window around its annotated $t_f$. We then extract the corresponding failure window according to its timestamp and feed it together with the successful window into the feature encoder to obtain high-dimensional representations. Among the candidates, we select the failure segment with the smallest feature distance to the successful segment. Algorithm~\ref{alg:stage2} summarizes one Stage 2 training step.

\begin{algorithm}[h]
\caption{Stage-2 Boundary-Aware Contrastive Training Step}
\label{alg:stage2}
\small
\begin{algorithmic}[1]
\Require Success batch $\mathcal{B}^{+}$; failure dataset $\mathcal{D}^{-}$; pretrained encoder $\phi_\theta$ and predictor $g_\psi$; margin $m$; loss weight $\lambda_c$; match tolerance $\Delta$; window size $|\mathcal{W}|$
\Ensure Updated model parameters $\theta, \psi$

\State $\mathcal{L}_{\mathrm{pred}} \gets$ Compute prediction loss on $\mathcal{B}^{+}$ and sampled failure batch from $\mathcal{D}^{-}$ (masking past truncation)
\State $\mathcal{H} \gets \varnothing$ \Comment{Initialize hit set for matched pairs}

\ForAll{success sample $i \in \mathcal{B}^{+}$}
    \State Candidate set $\mathcal{C}_i \gets \big\{ \xi^{-} \in \mathcal{D}_{l_i}^{-} \;\big|\; |t - t_f| \le \Delta \big\}$ \Comment{Filter candidates by progress anchor}
    \If{$\mathcal{C}_i \neq \varnothing$}
        \State $\xi_i^{-} \gets \operatorname*{arg\,min}_{\xi^{-} \in \mathcal{C}_i} \frac{1}{|\mathcal{W}|} \sum_{\tau \in \mathcal{W}} \big\| \phi_\theta(\mathcal{W}_t(\xi_i^{+})) - \phi_\theta(\mathcal{W}_t(\xi^{-})) \big\|_2^2$ \Comment{Feature matching,~\Eqref{eq:match}}
        \State $\mathcal{H} \gets \mathcal{H} \cup \{(i, \xi_i^{-})\}$
    \EndIf
\EndFor

\State Compute predicted latent sequences $z_{t+\mathcal{W}}^{+}$ and $z_{t+\mathcal{W}}^{-}$ for all pairs $(i, \xi_i^{-}) \in \mathcal{H}$
\State $\mathcal{L}_{\mathrm{con}} \gets 0$

\ForAll{matched pair $(i, \xi_i^{-}) \in \mathcal{H}$}
    \State $\alpha \gets \operatorname{softmax}\Big( \frac{1}{L H_a Q}\sum_{l=1}^L \sum_{h=1}^{H_a} \sum_{q=1}^Q A^{(l,h,q)} \Big)$ \Comment{Extract temporal salience profile, detached via $\operatorname{sg}(\cdot)$}
    \State $u \gets \frac{1}{|\mathcal{W}|} \sum_{\tau \in \mathcal{W}} z_{t+\tau}^{+}$ \Comment{Compute trajectory summary descriptor}
    \State $g \gets \sigma(W_g u)$ \Comment{Compute dynamic channel mask,~\Eqref{eq:gating_def}}
    \State $d_m \gets \sum_{\tau \in \mathcal{W}} \alpha_\tau \cdot \big\| g \odot (z_{t+\tau}^{+} - z_{t+\tau}^{-}) \big\|_F^2$ \Comment{Weighted distance,~\Eqref{eq:weighted_metric}}
    \State $\mathcal{L}_{\mathrm{con}} \gets \mathcal{L}_{\mathrm{con}} + \big[ m - d_m \big]_{+}$ \Comment{Hinge loss,~\Eqref{eq:weighted_metric}}
\EndFor

\State $\mathcal{L}_{\mathrm{con}} \gets \frac{1}{|\mathcal{H}|} \mathcal{L}_{\mathrm{con}}$
\State \Return Total fine-tuning loss $\mathcal{L}_{\mathrm{ft}} = \mathcal{L}_{\mathrm{pred}} + \lambda_c \, \mathcal{L}_{\mathrm{con}}$
\end{algorithmic}
\end{algorithm}

\section{Q1 Probing Protocol}
\label{app:probe}
This section specifies exactly how the separability analysis of Q1/Q2 (Fig.~\ref{fig:vis_sep}, Tab.~\ref{tab:separability}) is constructed, so that every reported number can be reproduced from the raw checkpoints.

\paragraph{Evaluation set.}
We evaluate the probe on a held-out split that is excluded from all training mixtures. It contains 109 failure episodes from LIBERO-100, LIBERO-Object, LIBERO-Spatial, and LIBERO-Goal, each with a verified failure timestamp $t_f$, together with 100 task-matched successful episodes. We use the same preprocessing pipeline as the world model, including video decoding, resizing, and normalization. Each probe input is a 9-frame window consisting of 8 context frames and 1 target frame, together with the 8 executed $\Delta$qpos actions.

\paragraph{Window sampling.}
We sample 16 windows from each failure episode: 8 uniformly over task progress and 8 concentrated in the decision-critical region $p\in[0.7,0.98]$ relative to the annotated failure point. Successful episodes are sampled using the corresponding progress grid. For failure episodes, we additionally use the relative time $\tau=t-t_f$ to identify the temporal position of each window, which can be mapped to the corresponding task progress $p$.

\paragraph{Latent readout.}
Each window is passed through the world model to obtain its predicted future latent $z$ via a forward pass (~\Eqref{eq:predict}). The resulting features are z-normalized within each task, reduced to 64 dimensions with PCA, and classified using logistic regression with episode-level GroupKFold(5). Thus, no window from an episode is used to train the probe that evaluates that episode. The resulting out-of-sample predictions are z-normalized within each task and used as the \emph{failure score}. We report both window-level scores and episode-level mean scores.

\paragraph{Phase-wise analysis.}
To examine when failure information becomes separable, we compute AUC separately in four temporal phases: early ($p<0.35$), mid ($0.35\le p<0.70$), near-failure ($p\ge0.70$), and post-failure ($\tau\ge0$). As a control, we also distinguish early and late windows within successful episodes only. This control remains essentially unchanged after Stage~2 ($0.87\rightarrow0.86$), indicating that the observed improvement is not explained by generic task-progress encoding.

\paragraph{Boundary-score analysis.}
We further test whether the learned boundary generalizes beyond the failure region used for training. The classifier is trained only on near-failure windows from the final $7.5$,s before failure ($\tau\in[-150,0)$ at 20~fps), together with near-goal windows from successful episodes. It is then applied to all windows of the held-out episodes, including post-failure windows that are never used for training. We report the AUC on near-failure windows, the AUC on unseen post-failure windows, and the fraction of failure episodes whose mean post-failure score exceeds the median score of successful episodes. The complete evaluation procedure is summarized in Algorithm~\ref{alg:probe}.

\begin{algorithm}[h]
\caption{Episode-Held-Out Probing Protocol}
\label{alg:probe}
\small
\begin{algorithmic}[1]
\Require
windows $\mathcal{W}_{\mathrm{probe}}=\{(\hat{z}_i,y_i,\Delta t_i,p_i,e_i,l_i)\}_{i=1}^{N}$, where
$\hat{z}_i$ is the predicted future latent of window $i$,
$y_i\in\{0,1\}$ is its outcome label
($1$: success, $0$: failure),
$\Delta t_i=t_i-t_f$ is its temporal offset from failure onset,
$p_i$ is task progress,
$e_i$ is its episode ID,
$l_i$ is its task ID; number of folds $K=5$
\\
\Require
$K$ episode-disjoint folds
$\{(\mathcal{I}^{(k)}_{\mathrm{tr}},\mathcal{I}^{(k)}_{\mathrm{te}})\}_{k=1}^{K}$
obtained by GroupKFold grouped by episode ID $e$
\\[1mm]

\State z-normalize $\hat{z}_i$ independently within each task $l_i$
\For{$k=1,\ldots,K$}
    \State fit PCA with 64 components on
    $\{\hat{z}_i:i\in\mathcal{I}^{(k)}_{\mathrm{tr}}\}$
    \State project training and test latents using the fitted PCA
    \State fit a logistic regression readout on the projected training windows
    \State $s_i^{(k)}\gets$ logistic-regression decision value
    for all $i\in\mathcal{I}^{(k)}_{\mathrm{te}}$
\EndFor

\State z-normalize $s_i$ independently within each task $l_i$
\State \Return task-normalized $s_i$ as the \emph{failure score}
\State $\bar{s}_e\gets
\frac{1}{|\mathcal{W}_e|}
\sum_{i\in\mathcal{W}_e}s_i$
for each episode $e$,
where $\mathcal{W}_e=\{i:e_i=e\}$
\Comment{episode-level score}

\State compute window-level ROC-AUC from $(s_i,y_i)$
\State compute episode-level ROC-AUC from
$(\bar{s}_e,\bar{y}_e)$ using the corresponding episode outcomes
\State compute phase-wise ROC-AUC within progress-matched bands
\State compute the success-only early-vs-late control

\State \textbf{Boundary probing:}
restrict probe-training windows to
$\{\Delta t_i\in[-150,0)\}\cup
\{y_i=1,\ p_i\geq0.7\}$
\State repeat the same episode-held-out PCA and logistic fitting procedure
using only the restricted training windows
\State apply each fitted readout to \emph{all} windows in its held-out episodes,
including windows after failure onset
\State compute near-failure and unseen post-failure ROC-AUC
\State \Return boundary scores and corresponding evaluation metrics
\end{algorithmic}
\end{algorithm}

\section{Additional Ablations}
\label{app:ablation}

\subsection{Separability of Failure and Success in Single In-Distribution Tasks}

\begin{figure*}[t]
\centering
\includegraphics[width=\textwidth]{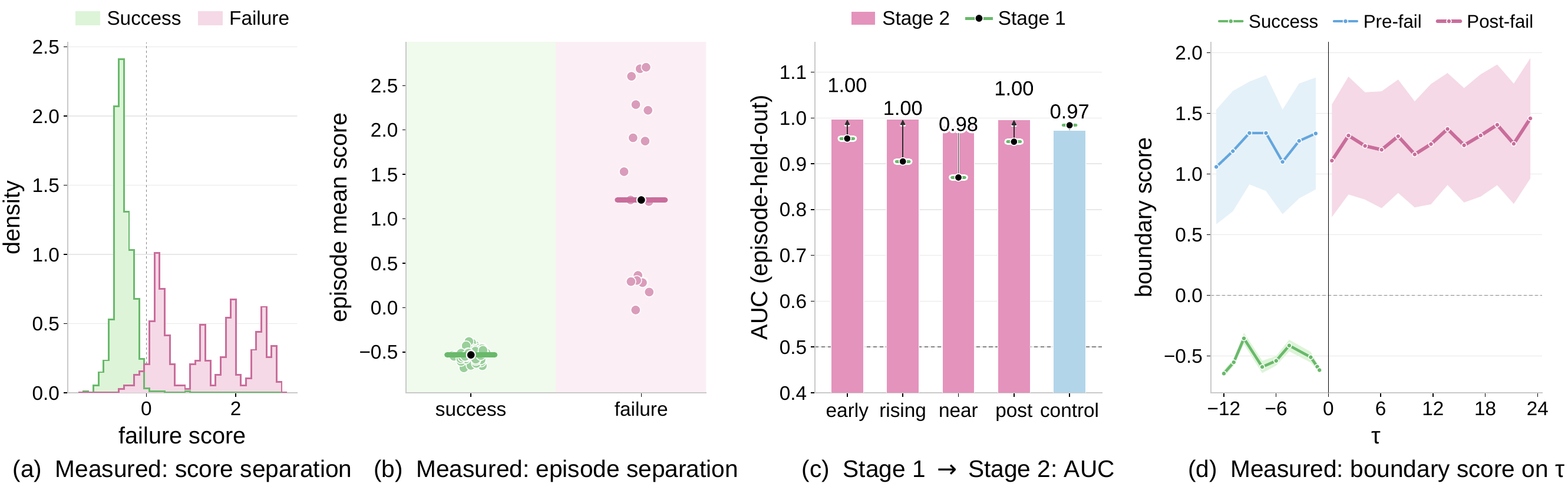}
\vspace{-5mm}
\caption{\textbf{Success--failure separability on RoboTwin \texttt{stack\_blocks\_three}.} (a) Single-window failure-score distributions. (b) Episode mean scores. (c) Stage-wise AUC. (d) Boundary score swept along the time offset $\tau$ from the expected task end.}
\label{fig:stack3_sep}
\end{figure*}

The readout utilizes linear probes (logistic regression on world-model embeddings) trained and evaluated via episode-held-out cross-validation to ensure zero data leakage across episodes on RoboTwin: 22 successful rollouts with VLM-annotated subtask timestamps versus 20 timeout failures (40\,s each, 14-DoF bimanual control). The primary probe emits a signed \textcolor{red}{\emph{failure score}} (positive indicates failure). Single-window scores yield near-disjoint distributions (Fig.~\ref{fig:stack3_sep}a), while their per-episode average (\textcolor{red}{\emph{episode mean score}}) confirms trajectory-level separation beyond local window noise, ranking all 17 failing trajectories above all 92 successful ones (Fig.~\ref{fig:stack3_sep}b). Quantitatively, episode-level AUC reaches $1.000$ with window-level AUC at $0.993$ under an EER of $3.2\%$. Separability remains uniform across execution progress: $0.997$, $0.997$, and $0.982$ over early, rising, and near-completion stage windows, and $0.996$ on post-completion windows, compared to $0.972$ for the success-only progress control. 

To characterize when the failure signature emerges relative to the expected task end, we train a second, \emph{phase-matched} probe to derive a \textcolor{red}{\emph{boundary score}}. Intuitively, this score measures whether a window resembles the moments immediately preceding an error (positive) or the winding-down phase of a success (negative). Formally, the probe is trained on late-stage failure windows ($\tau \in [-150, 0)$\,frames ) and late success windows ($p \ge 0.70$), thereby decoupling elapsed time from classification. Projecting all windows, including the unseen post-boundary segment ($\tau \ge 0$), onto this boundary score reveals that successful trajectories remain negative and deepen toward completion, whereas failing ones are already strongly positive well before the boundary and remain flat through the timeout horizon without decaying back toward the success side (Fig.~\ref{fig:stack3_sep}d). Before the boundary, the probe separates impending failure from late success at $0.98$ AUC; beyond it, the signature persists on held-out post-boundary windows ($0.996$ AUC, with every failing episode lying above the success median). Together, these results indicate that the failure-rich world model encodes a persistent, outcome-decisive failure signature for this task family, present from early execution rather than emerging only at a discrete error onset.

\subsection{Contrastive-Negative and Weighting Variants}

\definecolor{tableheader}{RGB}{240, 244, 248} 
\definecolor{tablehighlight}{RGB}{245, 247, 250} 

\begin{wraptable}{r}{0.52\textwidth}
\centering
\setlength{\tabcolsep}{3pt}
\setlength{\abovecaptionskip}{0pt}
\caption{\textbf{Stage-2 design choices.} Policy success rates (\%) on RoboTwin (clean/randomized placements) and LIBERO with one Stage-2 component replaced at a time; metric variants replace $d_m$ in both the contrastive loss and the Stage-3 reward.}
\label{tab:contrastive_variants}
\resizebox{\linewidth}{!}{%
\begin{tabular}{lcc}
\toprule
Variant & RoboTwin & LIBERO \\
\midrule
\rowcolor{tableheader}\multicolumn{3}{l}{\emph{Counterpart selection (separated with $d_m$)}} \\
Random opposite-outcome & 78.3/77.1 & 95.5 \\
Progress-anchored, random pick & 84.7/84.2 & 98.0 \\
\addlinespace
\rowcolor{tableheader}\multicolumn{3}{l}{\emph{Separation metric (matched counterparts)}} \\
Uniform $\alpha$ & 85.0/84.6 & 98.1 \\
$g \equiv \mathbf{1}$ & 85.3/85.0 & 98.2 \\
Plain $\ell_2$ (uniform $\alpha$, $g \equiv \mathbf{1}$) & 84.6/84.1 & 98.0 \\
\addlinespace
\rowcolor{tablehighlight}\textbf{Full (anchored $+$ nearest, $d_m$)} & \textbf{85.7/86.1} & \textbf{98.4} \\
\bottomrule
\end{tabular}}
\end{wraptable}

Tab.~\ref{tab:contrastive_variants} isolates the two core design choices of Stage~2, demonstrating that both are critical. Counterpart selection plays a dominant role: replacing the progress-anchored nearest-candidate rule with random opposite-outcome failures of the same task costs $7.4$ points on RoboTwin and $2.9$ points on LIBERO. Because random pairs differ in scene layout and execution progress, the contrastive gradient is wasted separating incidental cues rather than outcome-decisive ones. The progress anchor alone recovers most of this gap ($84.7/84.2$), confirming that task-phase alignment is the primary filter, while feature-nearest selection contributes the final point by concentrating supervision on visually subtle discrepancies. The separation metric contributes at a comparable magnitude: with matched counterparts but a plain $\ell_2$ metric, performance falls to $84.6/84.1$ (essentially the anchored-random level), as uninformative frames and static channels dilute the separation gradient and, through the shared Stage-3 reward, blunt the penalty at the decisive phase. Between the two weighting components, temporal salience $\alpha$ contributes more ($85.0/84.6$ without it versus $85.3/85.0$ without the channel gate $g$), consistent with outcome-decisive evidence concentrating in a few execution-critical steps. LIBERO margins are smaller throughout, reflecting the suite's near-saturation regime where RoboTwin's randomized placements expose robustness differences more sensitively.

\subsection{Hyper-parameter Sensitivity}
\label{app:sensitivity}

\begin{figure}[h]
\centering
\includegraphics[width=\textwidth]{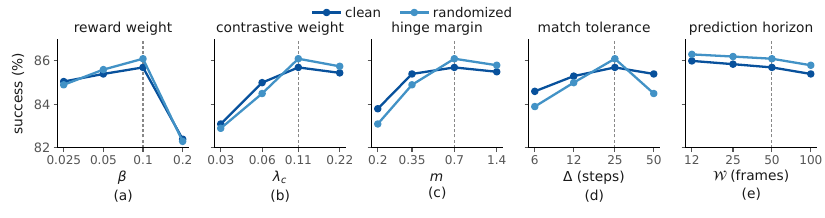}
\caption{\textbf{Hyper-parameter sensitivity.} Success rate on RoboTwin (50 episodes per task, clean and randomized splits) as a function of reward weight $\beta$, contrastive weight $\lambda_c$, hinge margin $m$, match tolerance $\Delta$, and prediction horizon $\mathcal{W}$, with all other hyper-parameters fixed at defaults (ringed marker, dashed line). The two loss weights show opposite asymmetries: under-weighting the boundary reward ($\beta$) is nearly harmless, whereas over-weighting it reduces performance by $3.3$ points; conversely, shrinking $\lambda_c$ removes Stage~2 entirely. The hinge margin $m$ acts as a threshold, saturating beyond $0.35$. Match tolerance $\Delta$ is comparatively flat, losing $2.2$ points on the randomized split at $\Delta{=}6$, where narrow windows fail to retrieve counterparts and starve the contrastive loss, while excessively wide windows mix execution phases. Horizon $\mathcal{W}$ declines slowly from $\mathcal{W}{=}12$ to $100$ ($-0.6$ points) because distant frames carry less decisive evidence; we set $\mathcal{W}{=}50$ to align with the policy's action chunk.}
\label{fig:sensitivity}
\end{figure}

Fig.~\ref{fig:sensitivity} sweeps each hyper-parameter of the three-stage objective over four log-spaced points around its default while holding all others fixed, reporting success on both RoboTwin splits. The five responses exhibit distinct behaviors. The two loss weights break symmetry in opposite directions: the Stage-3 reward weight $\beta$ is nearly inert on the low side ($85.05$ at $\beta{=}0.025$, within half a point of default) but drops by $3.3$ points when doubled to $0.2$, as the reward term overwhelms the imitation objective and the policy trades task completion for boundary avoidance. The contrastive weight $\lambda_c$ shows the converse, dropping to $83.1$ when shrunk below $0.06$ (an under-weighted Stage~2 contributes minimal gradient, effectively reverting to Stage~1 pretraining) while remaining within $0.3$ points of the peak when doubled. The hinge margin $m$ behaves as a threshold rather than a continuous dial: any value beyond $0.35$ lies on a plateau ($85.4$--$85.7$), making the default $0.7$ sit safely in the interior region. Match tolerance $\Delta$ produces the flattest curve on the clean split ($84.6$--$85.7$) and is only mildly steeper under randomized placements, where a $\Delta{=}6$ window occasionally fails to retrieve a valid counterpart ($-2.2$ points). Finally, prediction horizon $\mathcal{W}$ declines slowly and monotonically from the smallest window tested (dropping $0.6$ points from $\mathcal{W}{=}12$ to $100$): frames far ahead of the boundary carry progressively less outcome-decisive evidence, so we retain $\mathcal{W}{=}50$ not for accuracy gains but to align with the policy's action chunk length. Overall, no hyper-parameter requires sensitive tuning beyond avoiding the two extreme cliffs ($\beta$ too large or $\lambda_c$ too small), as all default values reside within broad, stable regions.

\subsection{Spatio-Temporal Factorization in $d_m$}

\begin{figure}[h]
\centering
\includegraphics[width=\textwidth]{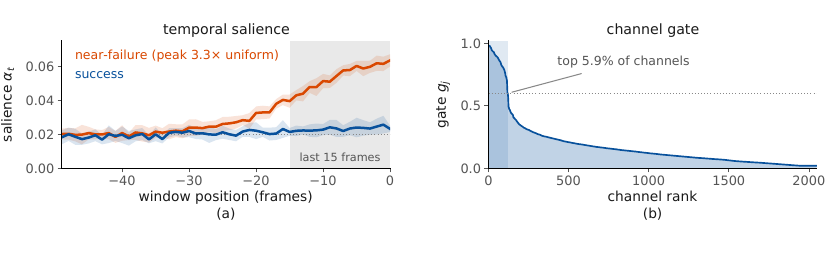}
\caption{\textbf{Learned weighting inside $d_m$.} (a)~Temporal salience $\alpha_t$ over the 50-frame context window. For near-failure windows, attention concentrates on the final $\sim 15$ frames preceding the boundary, whereas success windows remain nearly uniform, demonstrating that the model learns \emph{when} to attend. (b)~Sorted channel gate $g_j$. The gate maintains a decisive core of $\sim 6\%$ of the 2048 feature channels near saturation while suppressing the rest, indicating that failure evidence resides in a low-dimensional subspace rather than being distributed across the representation.}
\label{fig:alpha_g}
\end{figure}

Fig.~\ref{fig:alpha_g} examines what the two learned weightings inside $d_m$ actually encode. Panel~(a) traces the temporal salience $\alpha_t$ over the 50-frame context window. Because $\alpha$ is derived from the predictor's cross-attention maps and detached via a stop-gradient operation (Eq.~\ref{eq:gating_def}), its structure cannot be an artifact of the contrastive objective. On near-failure windows, $\alpha_t$ rises smoothly through the final $\sim 15$ frames before the boundary, peaking at $3.3\times$ the uniform baseline $1/|\mathcal{W}|$, whereas on success windows it remains within $1.2\times$ of uniform throughout. The world model thus autonomously identifies \emph{when} outcomes are decided without explicit supervision, while the flat success profile rules out a generic recency bias. Panel~(b) sorts the channel gate $g_j$, revealing that a decisive core of $\sim 5.9\%$ of the 2048 predicted-feature channels sits above the saturation threshold ($g_j > 0.6$) while the remainder is heavily suppressed. This indicates that failure evidence concentrates in a low-dimensional subspace of the representation rather than spreading across it. 

These two weight distributions explain the ablation pattern in Tab.~\ref{tab:contrastive_variants}. Removing $\alpha$ proves to be the costlier modification (dropping to $85.0/84.6$ with uniform temporal weighting) because outcome-decisive frames are sparse, making simple averaging over all 50 frames highly dilutive. Conversely, removing $g$ leaves correctly timed but unfiltered channels ($85.3/85.0$), whereas plain $\ell_2$ eliminates both spatial and temporal concentrations, falling back to the anchored-random baseline ($84.6/84.1$). This dual spatio-temporal factorization also enables $d_m$ to double as the Stage-3 reward: a metric that selectively highlights outcome-decisive frames and channels naturally serves as a sharp, responsive failure signal during deployment.

\section{More Visualizations}

\begin{figure*}[h]
\centering
\includegraphics[width=\textwidth]{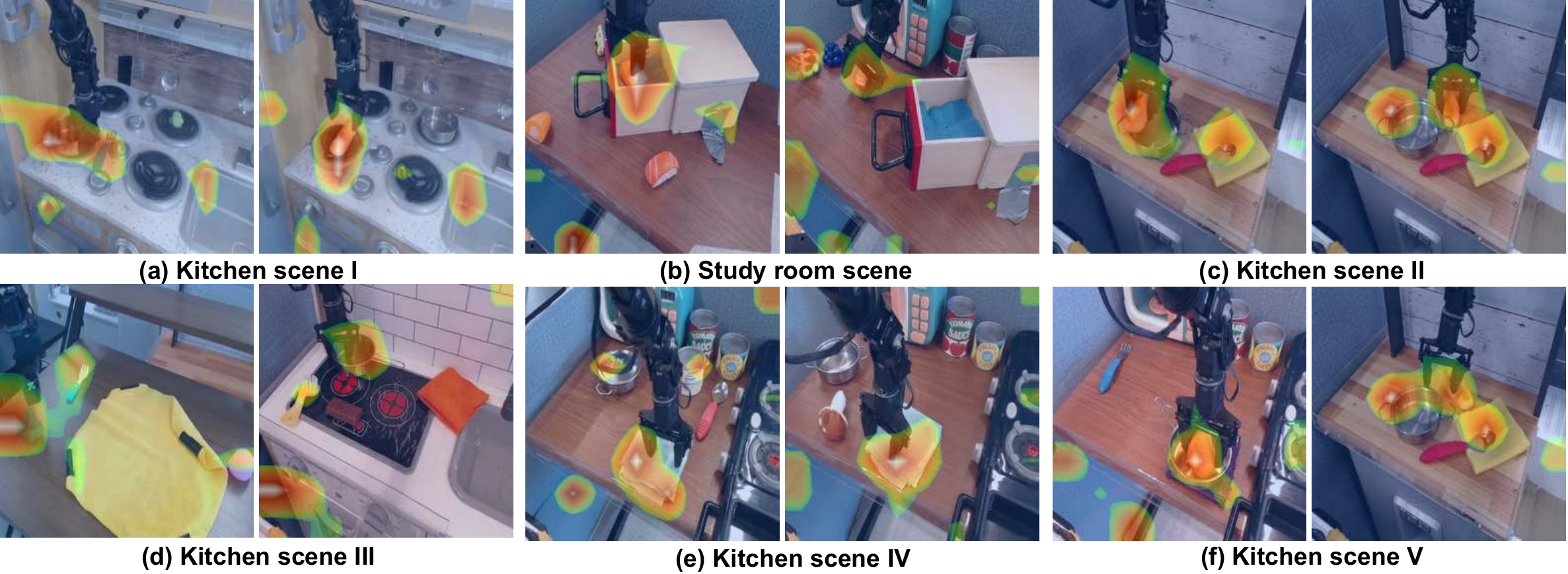}
\vspace{-5mm}
\setlength{\belowcaptionskip}{0pt}
\caption{Attention weight matrix of latent action tokens to image tokens on SimplerEnv.}
\vspace{-4mm}
\label{fig:simpler_env_attn}
\end{figure*}

\begin{figure*}[h]
\centering
\includegraphics[width=\textwidth]{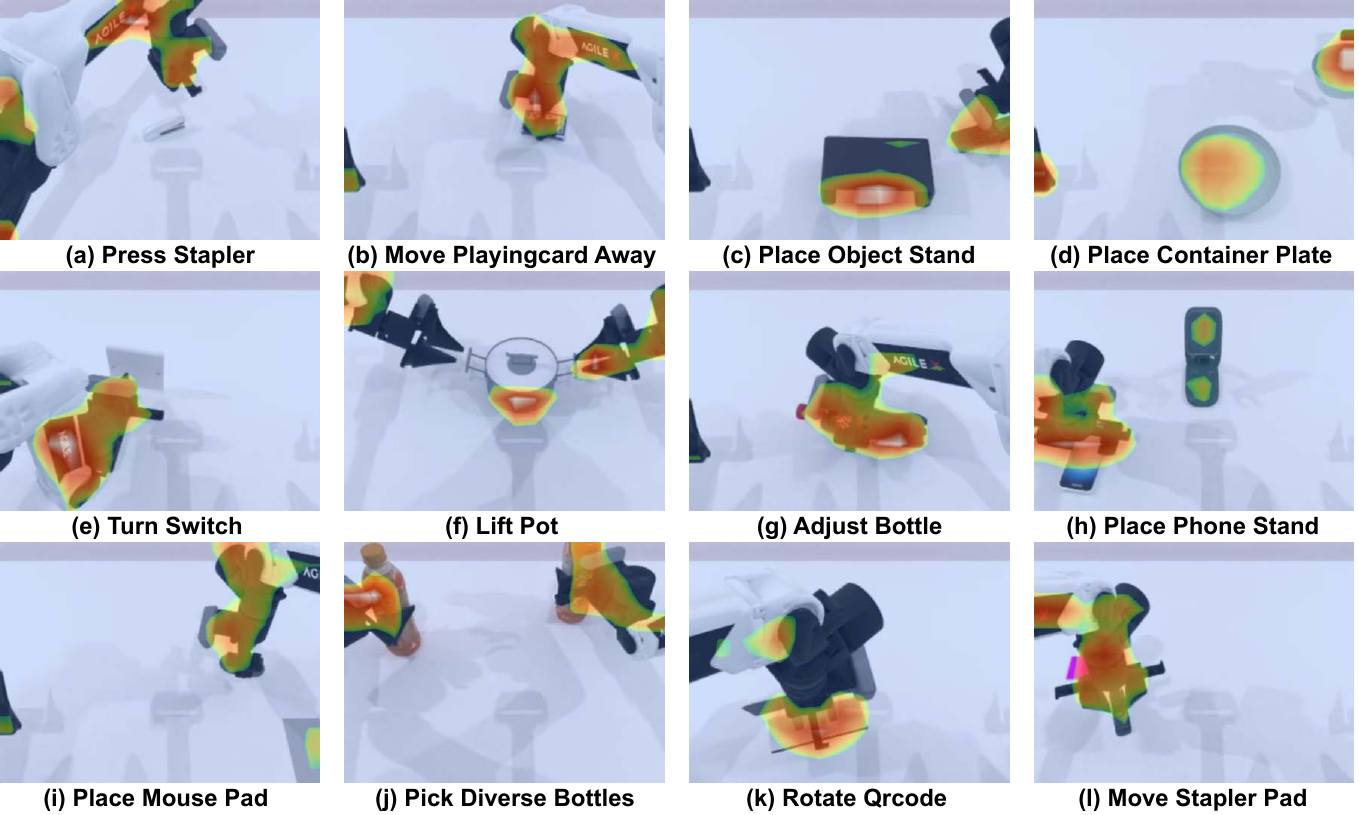}
\vspace{-5mm}
\setlength{\belowcaptionskip}{0pt}
\caption{Attention weight matrix of latent action tokens to image tokens on Robotwin.}
\vspace{-4mm}
\label{fig:robotwin_attn}
\end{figure*}

To intuitively understand how the proposed model bridges visual perception and action prediction, we visualize the attention weight matrices from the latent action tokens to the input image tokens across SimplerEnv (Fig.~\ref{fig:simpler_env_attn}) and Robotwin (Fig.~\ref{fig:robotwin_attn}). As depicted in Fig.~\ref{fig:simpler_env_attn}, across diverse domains---including multiple kitchen environments (Kitchen scene I--V) and the study room scene---the visual attention maps consistently focus on task-critical regions. Specifically, high-attention activations (highlighted in red and orange) are strictly concentrated on the robotic end-effectors, target manipulation objects (e.g., handles, bowls, spoons, and buttons), and their immediate interaction boundaries. Irrelevant background elements (such as wall tiles, countertops, and distant background clutter) receive minimal attention, demonstrating the model's strong immunity to visual distractor noise.

Similarly, Fig.~\ref{fig:robotwin_attn} showcases the attention distributions across 12 distinct manipulation tasks on the Robotwin benchmark. The heatmaps reveal precise spatial grounding aligned with specific task semantics:
\begin{enumerate}
    \item \textbf{Tool \& Lever Operations:} In tasks like Press Stapler (a), Turn Switch (e), and Rotate Qrcode (k), the attention peaks precisely at the contact points between the gripper and the functional parts of the tools.
    \item \textbf{Object Placement \& Alignment:} For placement tasks such as Place Object Stand (c), Place Container Plate (d), and Place Mouse Pad (i), high-weight tokens track both the held object and the target placement region, indicating that the latent action representations dynamically encode spatial relationships.
    \item \textbf{Bimanual \& Fine-Grained Manipulation:} In complex operations like Lift Pot (f) and Pick Diverse Bottles (j), the model simultaneously activates regions corresponding to both robotic arms and the object handles, confirming robust task-level coordination.
\end{enumerate}

Overall, these visualization results demonstrate that the learned world model effectively grounds action latent representations into semantic and physical contact regions, providing high interpretability and explaining its strong generalization across diverse scenes and tasks.

\subsection{Implications for Recovery Performance}
\label{app:recovery_performance_discussion}

Currently, Stage 2 training relies on finely annotated data and requires collecting failure trajectories using the model, which makes data construction time-consuming. In future work, we will focus on addressing this limitation.

\end{document}